\documentclass[11pt]{article}

\usepackage[final]{acl}

\usepackage{times}
\usepackage{latexsym}

\usepackage{amsmath,amsfonts,bm}

\def\eqref#1{equation~\ref{#1}}

\def\1{\bm{1}}

\DeclareMathAlphabet{\mathsfit}{\encodingdefault}{\sfdefault}{m}{sl}
\SetMathAlphabet{\mathsfit}{bold}{\encodingdefault}{\sfdefault}{bx}{n}

\newcommand{\E}{\mathbb{E}}

\usepackage[T1]{fontenc}

\usepackage[utf8]{inputenc}

\usepackage{microtype}

\usepackage{inconsolata}

\usepackage{graphicx}

\usepackage{hyperref}       
\usepackage{url}            
\usepackage{booktabs}       
\usepackage{amsfonts}       
\usepackage{nicefrac}       
\usepackage{xcolor}         

\usepackage{xspace}
\usepackage{amsmath}
\usepackage{cleveref}
\usepackage{algorithm}
\usepackage{algorithmic}
\usepackage{wrapfig}
\usepackage{bbm}
\usepackage{tabularx}
\usepackage{multirow}
\usepackage{graphicx}
\usepackage{array}
\usepackage{subcaption}
\usepackage{enumitem}
\usepackage[export]{adjustbox}   

\newcommand{\answer}{a}
\newcommand{\question}{Q}
\newcommand{\rationale}{E}
\newcommand{\image}{I}
\newcommand{\answerset}{\mathcal{A}}
\newcommand{\quality}{\varphi}
\newcommand{\model}[1]{m_{#1}}
\newcommand{\qas}{\texttt{QA}\rightarrow\texttt{S}}
\newcommand{\nli}{\texttt{NLI}}
\newcommand{\qgen}{\texttt{QGen}}

\newcommand{\verif}{\texttt{Verif}}
\newcommand{\visualfidelity}{\mathrm{VF}}
\newcommand{\contrastiveness}{\mathrm{Contr.}}
\newcommand{\acc}{\text{Acc}}
\newcommand{\disc}{\text{Disc}}

\newcommand{\avg}{\mathrm{avg}}
\newcommand{\product}{\mathrm{Prod}}

\definecolor{darkblue}{rgb}{0, 0, 0.5}
\hypersetup{colorlinks=true, citecolor=darkblue, linkcolor=darkblue, urlcolor=darkblue}

\renewcommand{\cite}[1]{\citep{#1}}


\title{Believing without Seeing: Quality Scores for Contextualizing Vision-Language Model Explanations}

\author{
    Keyu He\textsuperscript{1,2} \quad Tejas Srinivasan\textsuperscript{1} \quad Brihi Joshi\textsuperscript{1} \\
    \textbf{Xiang Ren\textsuperscript{1}} \quad \textbf{Jesse Thomason\textsuperscript{1}} \quad \textbf{Swabha Swayamdipta\textsuperscript{1}} \\
    \textsuperscript{1}University of Southern California\quad
    \textsuperscript{2}Carnegie Mellon University 
    \\
    \texttt{keyuhe@cmu.edu}
}

\newcommand{\llava}{LLaVA-v1.5-7B\xspace}
\newcommand{\qwen}{Qwen2.5-VL-7B\xspace}
\newcommand{\gemma}{Gemma-3n-E4B\xspace}
\newcommand{\gpt}{GPT-4o\xspace}
\newcommand{\gptmini}{GPT-4o-mini\xspace}

\begin{document}
\maketitle

\begin{abstract}
When people query Vision-Language Models (VLMs) but cannot see the accompanying visual context (e.g. for blind and low-vision users), augmenting VLM predictions with natural language explanations can signal which model predictions are reliable. 
However, prior work has found that explanations can easily convince users that inaccurate VLM predictions are correct. 
To remedy undesirable overreliance on VLM predictions, we propose evaluating two complementary qualities of VLM-generated explanations via two quality scoring functions.
We propose \emph{Visual Fidelity}, which captures how faithful an explanation is to the visual context, and \emph{Contrastiveness}, which captures how well the explanation identifies visual details that distinguish the model's prediction from plausible alternatives.
On the A-OKVQA, VizWiz, and MMMU-Pro tasks, these quality scoring functions are better calibrated with model correctness than existing explanation qualities. 
We conduct a user study in which participants have to decide whether a VLM prediction is accurate without viewing its visual context. 
We observe that showing our quality scores alongside VLM explanations improves participants' accuracy at predicting VLM correctness by 11.1\%, including a 15.4\% reduction in the rate of falsely believing incorrect predictions. 
These findings highlight the utility of explanation quality scores in fostering appropriate reliance on VLM predictions.

\end{abstract}

\section{Introduction}
\label{sec:intro}
Vision-Language Models (VLMs) are being deployed in applications where users who do not have access to the VLM's visual context; for example, in assisting blind and low-vision individuals~\cite{ huh2024longform,an2025lvlmsautomaticmetricscapture, kim2025guidedogrealworldegocentricmultimodal}, autonomous multimodal digital agents~\cite{koh-etal-2024-visualwebarena}, and in human-robot collaboration \cite{lukin-etal-2018-scoutbot}.
However, despite recent advances in VLM capabilities, they often exhibit unreliable behavior, including hallucinating visual details~\cite{li2023evaluating, gunjal2024detecting} and making overconfident predictions~\cite{valdenegro2024overconfidence}.
In scenarios where users cannot directly observe the visual context of the VLM, it becomes imperative to enable users to accurately trust model outputs.
How can we provide adequate context for users to establish appropriate reliance on model predictions and explanations?

\begin{figure*}[t]
    \centering
    \vspace{-10pt}
    \includegraphics[width=\linewidth]{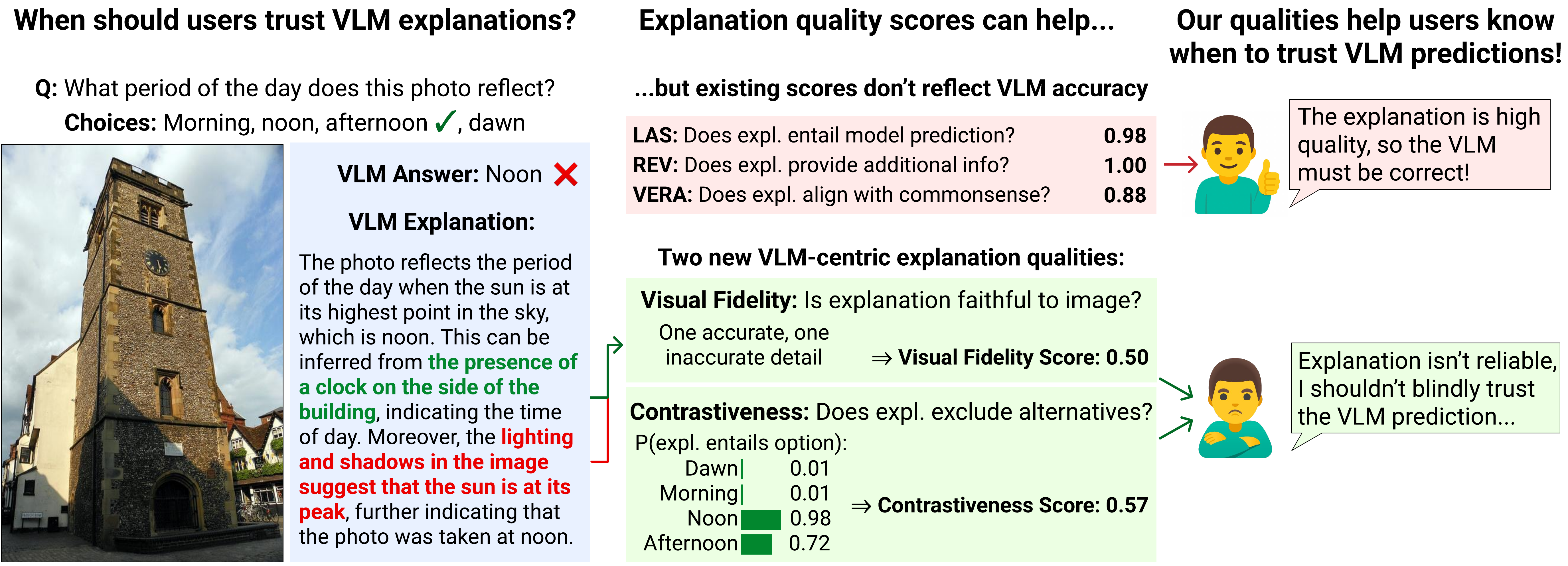}
    \vspace{-20pt}
    \caption{VLM explanations that sound plausible can mislead users.
    Contextualizing VLM explanations with quality scores may help users know when to rely, but existing explanation qualities are not calibrated with VLM prediction accuracy.
    We propose Visual Fidelity and Contrastiveness to contextualize these explanations, which are better calibrated with correctness, and also help users make better decisions about when to believe VLM predictions.
    }
    \label{fig:fig1}
\end{figure*}

Prior work has explored the utility of model explanations to support user decision making~\cite{wang2021explanations, bansal2021does}. 
However, natural language explanations can be misleading for inaccurate model predictions~\cite{joshi-etal-2023-machine,chaleshtori2024evaluating, si2024largelanguagemodelshelp, sieker2024illusion}. 
In \autoref{fig:fig1}, we consider the question ``\emph{What period of the day does this photo reflect?}'' where a VLM (incorrectly) answers ``\emph{Noon}'' and generates a plausible explanation referencing cues like ``\emph{a clock on the building}'' or ``\emph{lighting and shadows}.'' 
This reasoning may appear highly convincing to a user without access to the image, even though the prediction is incorrect.
Users are particularly susceptible to \emph{overreliance} on explanations in scenarios with such information asymmetry.

Existing explanation metrics~\cite{hase2020leakage, chen2023rev, liu2023vera} are designed to evaluate explanations generated by text-only LLMs, and we find these are \emph{poorly calibrated} to VLM prediction accuracy. 
However, considering the explanation with respect to the accompanying visual context can reveal the reliability of the VLM prediction; for instance, identifying hallucinated details in the explanation (\emph{``shadows suggesting that the sun is at its peak''}) or that the explanation does not mention the critical detail for identifying the correct answer (the time on the clock).

We propose evaluating two new qualities of VLM-generated explanations: \emph{Visual Fidelity}---how faithfully a VLM explanation reflects the accompanying visual context---and \emph{Contrastiveness}---how well the explanation rules out alternative answers. 
We introduce training-free methods to estimate both these qualities without relying on ground-truth quality annotations.

In experiments with three popular VLMs and three visual reasoning tasks, A-OKVQA \cite{schwenk2022okvqa}, VizWiz \cite{gurari2018vizwiz}, and MMMU-Pro \cite{yue2025mmmuprorobustmultidisciplinemultimodal}, we find that our Visual Fidelity and Contrastiveness qualities better distinguish correct VLM predictions than do existing evaluation qualities~\cite{hase2020leakage, chen2023rev, liu2023vera}, 
and are also better calibrated with VLM correctness (\S\ref{sec:metric_experiments}).
We conduct several user studies in which participants evaluated VLM predictions without access to visual inputs (\S\ref{sec:user_studies}). 
We find that augmenting VLM explanations with our proposed qualities improves participants’ ability to distinguish between correct and incorrect VLM predictions.
In particular, showing the product of Visual Fidelity and Contrastiveness as a single quality score leads to an 11.1\% absolute improvement in user accuracy, along with a 15.4\% absolute reduction in the rate of users falsely believing inaccurate predictions. 
Finally, we show that presenting explanation qualities via natural language descriptions instead of scores further improves user decision making accuracy (\S\ref{subsec:userstudy_experiments}). 
Overall, our findings highlight the utility of evaluating the quality of VLM-generated explanations and communicating these qualities to users relying on VLM assistants.

\section{Quality Scores for VLM Explanations}
\label{sec:methods}
Visual reasoning is the task of answering a textual question about an image by drawing inferences from what is visually observed. 
A vision-language model (VLM) is given an input $x = (\image, \question) \in \mathcal{X}$ consisting of an image $\image$ and a question $\question$, and produces an answer $\answer \in \answerset$, a (closed or open) set of options. 
Additionally, the VLM generates a natural language explanation $\rationale$ for its prediction $\answer$. 
This explanation can be either a chain-of-thought rationale~\cite{wei2023chainofthoughtpromptingelicitsreasoning} that the VLM prediction is conditioned on $(\image \question \rightarrow \rationale \answer)$ or a post hoc justification generated after the VLM prediction $(\image \question \rightarrow \answer \rationale)$~\cite{wiegreffe2022measuringassociationlabelsfreetext}. 

We define an explanation quality scoring function $S_{\quality}(\rationale; \image, \question, \answer, \answerset)$\footnote{For brevity, we note arguments for $S$ only when relevant.} that maps a VLM-generated explanation $\rationale$ to a score in $[0,1]$ reflecting a specific quality $\quality$ of the reliability of an explanation.
Existing qualities for natural language explanations, primarily developed and evaluated in text-only settings, typically look at how well an explanation supports the model's prediction~\cite{hase2020leakage}, its informativeness~\cite{chen2023rev}, and its plausibility according to common sense priors~\cite{liu2023vera}. 
However, evaluating \textit{only} these qualities is insufficient for VLM-generated explanations since they do not consider the relationship between the explanation and the visual context. 
Further, we find that these explanation qualities are poorly calibrated with prediction accuracy (\S\ref{sec:metric_experiments}) and therefore not useful for a user deciding whether to trust the VLM's prediction or not.

To address these gaps, we introduce two \emph{novel trust cues}: \textbf{Visual Fidelity} and \textbf{Contrastiveness}. 
(Figure~\ref{fig:fig1}).

\subsection{Visual Fidelity}
\label{subsec:vf_metric}
Visual Fidelity quality captures how faithful the VLM's explanation is to the contents of the image. 
The explanation, typically spanning multiple sentences, may mention several details about the image in order to explain its prediction. 
However, it is possible that some of these details are hallucinated by the model~\cite{gunjal2024detecting}. 
For example, in \autoref{fig:fig1}, the VLM justifies its prediction by incorrectly claiming that the shadows indicate that the sun is at its peak. 
Such misleading details can sway users who cannot observe the VLM's visual context.
Evaluating the Visual Fidelity of an explanation could potentially alert users when important explanation details may be hallucinated.

\begin{algorithm}[t]
\caption{Evaluate Visual Fidelity}
\label{alg:visualfidelity}
\begin{algorithmic}[1]
\REQUIRE Image $\image$, question $\question$, model prediction $\answer$, explanation $\rationale$ \\
\ENSURE Visual Fidelity score $S_{\visualfidelity}(\rationale)$ 
\STATE $\mathcal{Q}^V = \{q^V_1, \dots, q^V_K\}
\leftarrow \model{\qgen}(\rationale, \question, \answer)$
\FORALL{$q^V_j \in \mathcal{Q}^V$}
  \STATE $a^V_j \leftarrow \model{\verif}(q^V_j, \image)$
\ENDFOR
\STATE $S_{\visualfidelity}(\rationale) \leftarrow \frac{\sum_{i=1}^{K} \mathbbm{1}\{a^V_i = \text{``yes''}\}}{K}$
\end{algorithmic}
\end{algorithm}

We measure the visual fidelity of an explanation $\rationale$ with respect to the image $\image$ by decomposing the explanation into a set of facts about the image and individually verifying each fact (Algorithm~\ref{alg:visualfidelity}).
Conceptually, this is an \emph{oracle} scoring function that would check each atomic claim against ground-truth visual Q\&A; we approximate it with an LLM-as-judge, using a verifier $\model{\verif}$ (\gpt by default), and we audit this approximation with expert annotations and verifier swaps at \Cref{sec:verifier_reliability}. 
Concretely, we prompt a question generation LLM $\model{\qgen}$ to generate a set of verification questions $\mathcal{Q}^V$ that confirm visual details that have been mentioned in the VLM-generated explanation.
The model has to generate questions such that answering ``yes'' confirms the presence of the detail in the image (see \Cref{tab:VF_prompt_1} for the full LLM prompt). 
The number of questions generated by $\model{\qgen}$ depends on the explanation and may vary across instances.

For each verification question  $q^V_i \in \mathcal{Q}^V$, a verifier VLM 
$\model{\verif}$ produces an answer $a^V_i \in \{\text{yes, no}\}$. 
The Visual Fidelity score $S_{\visualfidelity}(\rationale)$ is calculated as the fraction of questions for which the verifier $\model{\verif}$ answers ``yes'':
\begin{align*}
    S_{\visualfidelity}(\rationale) = \frac{\sum_{i=1}^{|\mathcal{Q}^V|} \mathbbm{1}\{a^V_i = \text{``yes''}\}}{|\mathcal{Q}^V|}
\end{align*}
A high Visual Fidelity score indicates that the rationale faithfully represents the contents of the image.

\subsection{Measuring Contrastiveness}
\label{subsec:contr_metric}
Contrastiveness captures whether the VLM's explanation mentions all relevant visual details to identify the correct answer. 
The evidences and reasoning in the explanation may not identify the visual details that correctly distinguish the correct answer from alternatives. 
For instance, the explanation in Figure~\ref{fig:fig1} fails to mention the time on the clock---the key visual cue for correctly answering the question. 
When such relevant cues are missed, the explanation may inadvertently support other alternative answers apart from the VLM's prediction. 
Evaluating the Contrastiveness of an explanation could potentially alert users to when the model explanation fails to identify relevant visual cues.

\begin{algorithm}[t]
\caption{Evaluate Contrastiveness}
\label{alg:contrastiveness}
\begin{algorithmic}[1]
\REQUIRE Question $\question$, explanation $\rationale$, set of possible answers $\answerset$, model prediction $\answer_0 \in \answerset$
\ENSURE Contrastiveness score $S_{\contrastiveness}(\rationale)$
\STATE \(P \leftarrow \text{Mask}(\rationale, \answerset)\)
\FORALL{\(a_j \in \answerset\)}
  \STATE \(h_j \leftarrow \model{\qas}(\question, a_j)\)
\ENDFOR
\STATE $S_{\contrastiveness}(\rationale) \leftarrow \frac{\textsc{Pr}_{\nli}(P \text{ entails } h_0)}{\sum_{a_j \in \answerset} \textsc{Pr}_{\nli}(P \text{ entails } h_j)}$
\end{algorithmic}
\end{algorithm}

However, evaluating whether all relevant visual details have been mentioned in the explanation is difficult if we do not know what the relevant visual cues are. 
For example, if the explanation never mentioned a clock, then we would not know that the time on the clock is the critical detail for predicting the correct answer. 
Therefore, we estimate sufficiency using a proxy measure, based on the insight that if the model did capture all relevant details then it would have eliminated other plausible alternatives. 
Specifically, for a task where a VLM has to predict an answer $\answer_0$ 
from a closed set of possible answers $\answerset$, we measure the Contrastiveness of an explanation $E$ by evaluating how strongly it entails the model prediction $\answer_0$ relative to the set of alternative answers $\answerset \backslash \{\answer_0\}$ (Algorithm~\ref{alg:contrastiveness}).

We begin by masking mentions of all answers $\answer \in \answerset$ in the explanation $\rationale$ to prevent label leakage from affecting the entailment model. 
Then, for each possible answer $\answer_j \in \answerset$, a paraphraser LLM $\model{\qas}$ paraphrases the question-answer pair $(\question, \answer_j)$ into a declarative sentence $h_j$. 
Finally, we calculate the probability that the masked explanation $P$ entails the hypothesis $h_j$ using an entailment model $\model{\nli}$.
The Contrastiveness $S_{\contrastiveness}(\rationale)$ score is computed as the relative entailment probability for the predicted hypothesis $h_0$, compared to the entailment probability over all possible hypotheses:
\begin{align*}
    S_{\contrastiveness}(\rationale) = \frac{\textsc{Pr}_{\nli}(P \text{ entails } h_0)}{\sum_{a_j \in \answerset} \textsc{Pr}_{\nli}(P \text{ entails } h_j)}.
\end{align*}
A high Contrastiveness score indicates the explanation identifies visual cues that not only support the prediction but also eliminate plausible alternatives.

\subsection{Combining Visual Fidelity and Contrastiveness}
\label{subsec:combined_metrics}
We study how complementary our scoring functions are by computing a single quality score that combines the two explanation quality scores $S_{\visualfidelity}(\rationale)$ and $S_{\contrastiveness}(\rationale)$. 
We evaluate three combinations: the average, product, and minimum of the Visual Fidelity and Contrastiveness quality scores.
\begin{align*}
    S_{\avg}(\rationale) = \frac{S_{\visualfidelity} + S_{\contrastiveness}}{2} ;\\ \quad S_{\product}(\rationale) = S_{\visualfidelity} \times S_{\contrastiveness} ;\\ \quad S_{\min} = \min(S_{\visualfidelity}, S_{\contrastiveness}).
\end{align*}

\section{Are Visual Fidelity and Contrastiveness Indicative of VLM Correctness?}
\label{sec:metric_experiments}

We evaluate quality scoring functions on their ability to indicate the accuracy of a VLM prediction.
\paragraph{Visual Reasoning Tasks.}
\label{subsec:vl-reasoning-task}
We evaluate our explanation qualities on three visual reasoning tasks: A-OKVQA~\cite{schwenk2022okvqa}---a multiple-choice VQA benchmark that requires reasoning over images using external knowledge and commonsense; VizWiz~\cite{gurari2018vizwiz}---an open-ended VQA task consisting of questions asked by blind and low-vision users about images they captured on their mobile phones; and MMMU-Pro~\cite{yue2025mmmuprorobustmultidisciplinemultimodal}---a massive multi-discipline multimodal understanding and reasoning benchmark.
We sample 500 questions from the validation set of each task. 
\Cref{subsec:create_subset_split} contains details about the tasks' pre-processing and evaluation. 
For VizWiz, which lacks a specified set of possible answers, we generate hard negatives using an LLM to enable Contrastiveness evaluation. These negatives are used for post-hoc evaluation and are not provided to the model during answer generation.\footnote{We discuss the specific challenges and results of this open-set evaluation in \Cref{sec:apx_hard_negatives} and validate the strategy via ablation in \Cref{sec:apx_hard_negatives_validation}.}

\begin{table*}[t]
\centering
\resizebox{\linewidth}{!}{
\begin{tabular}{@{}lrrrc|rrrc|rrrc|c@{}}
    \toprule
     & \multicolumn{13}{c}{\textbf{Discriminability (higher is better)}} \\ 
    \cmidrule(lr){2-14}
     & \multicolumn{4}{c}{A-OKVQA} & \multicolumn{4}{c}{VizWiz} & \multicolumn{4}{c}{MMMU-Pro} & \multirow{2}{*}{\textbf{\shortstack{Overall\\Avg}}} \\ 
    \cmidrule(lr){2-5} \cmidrule(lr){6-9} \cmidrule(lr){10-13}
     & LLaVA & Qwen2.5 & GPT-4o & Avg & LLaVA & Qwen2.5 & GPT-4o & Avg & LLaVA & Qwen2.5 & GPT-4o & Avg \\ 
    \midrule
    Simulatability & $0.106^{**\phantom{*}}$ & $0.247^{***}$ & $0.098^{\phantom{***}}$ & $0.150$ & $0.088^{*\phantom{**}}$ & $0.158^{**}$ & $-0.130^{*}$ & $0.039$ & $0.059^{\phantom{**}}$ & $0.195^{***}$ & $0.079^{**\phantom{*}}$ & $0.111$ & $0.100$\\
    Informativeness & $-0.018^{\phantom{***}}$ & $0.000^{\phantom{***}}$ & $0.069^{\phantom{***}}$ & $0.017$ & $0.093^{*\phantom{**}}$ & $-0.007^{\phantom{**}}$ & $-0.113^{\phantom{*}}$ & $-0.009\phantom{-}$ & $-0.008^{\phantom{**}}$ & $0.006^{\phantom{***}}$ & $0.011^{\phantom{***}}$ & $0.003$ & $0.004$\\
    Plausibility & $0.043^{**\phantom{*}}$ & $0.022^{\phantom{***}}$ & $0.028^{\phantom{***}}$ & $0.031$ & $0.004^{\phantom{***}}$ & $-0.020^{\phantom{**}}$ & $-0.002^{\phantom{*}}$ & $-0.006\phantom{-}$ & $0.006^{\phantom{**}}$ & $0.012^{\phantom{***}}$ & $0.004^{\phantom{***}}$ & $0.007$ & $0.011$\\
    \midrule
    Visual Fidelity & $0.181^{***}$ & $0.080^{***}$ & $0.085^{***}$ & $0.115$ & $0.240^{***}$ & $0.042^{*\phantom{*}}$ & $0.024^{\phantom{*}}$ & $\textbf{0.102}$ & $0.169^{**}$ & $0.038^{\phantom{***}}$ & $0.090^{***}$ & $0.099$ & $0.106$\\
    Contrastiveness & $0.243^{***}$ & $0.283^{***}$ & $0.248^{***}$ & $0.258$& $0.041^{\phantom{***}}$ & $0.054^{\phantom{**}}$ & $-0.097^{\phantom{*}}$ & $-0.001\phantom{-}$ & $0.047^{\phantom{**}}$ & $0.154^{***}$ & $0.101^{***}$ & $0.101$ & $0.119$\\
    \midrule
    Avg(VF, Contr.) & $0.212^{***}$ & $0.181^{***}$ & $0.166^{***}$ & $0.187$ & $0.141^{***}$ & $0.048^{\phantom{**}}$ & $-0.037^{\phantom{*}}$ & $0.051$ & $0.108^{**}$ & $0.096^{***}$ & $0.096^{***}$ & $0.100$ & $0.112$\\
    Prod(VF, Contr.) & $0.320^{***}$ & $0.315^{***}$ & $0.266^{***}$ & $\textbf{0.301}$& $0.168^{***}$ & $0.082^{\phantom{**}}$ & $-0.082^{\phantom{*}}$ & $\underline{0.056}$ & $0.090^{**}$ & $0.146^{***}$ & $0.143^{***}$ & $\textbf{0.126}$ & $\textbf{0.161}$\\
    Min(VF, Contr.) & $0.295^{***}$ & $0.298^{***}$ & $0.255^{***}$ & $\underline{0.283}$ & $0.165^{***}$ & $0.077^{\phantom{**}}$ & $-0.084^{\phantom{*}}$ & $0.053$ & $0.082^{**}$ & $0.127^{***}$ & $0.124^{***}$ & $\underline{0.111}$ & $\underline{0.149}$\\
    \bottomrule
\end{tabular}
}
\caption{Discriminability (Disc., higher is better) for different explanation quality scoring functions across three models and three visual reasoning tasks. Significance is evaluated using an unpaired t-test: $^{***}$ $p<0.001$, $^{**}$ $p<0.01$, $^{*}$ $p<0.05$. Averages are not evaluated for significance. Notably, VizWiz--\gpt exhibits significantly lower and predominantly negative discriminability scores across all metrics; see \Cref{sec:apx_vizwiz_negatives} for a discussion of the underlying causes.}
\label{tab:disc}
\end{table*}

\begin{table*}[t]
\centering
\resizebox{\linewidth}{!}{
\begin{tabular}{@{}lcccc|cccc|cccc|c@{}}
    \toprule
     &  \multicolumn{13}{c}{\textbf{Expected Calibration Error (lower is better)}}\\
    \cmidrule(lr){2-14}
     & \multicolumn{4}{c}{A-OKVQA} & \multicolumn{4}{c}{VizWiz} & \multicolumn{4}{c}{MMMU-Pro} & \multirow{2}{*}{\textbf{\shortstack{Overall\\Avg}}} \\ 
    \cmidrule(lr){2-5} \cmidrule(lr){6-9} \cmidrule(lr){10-13}
     & LLaVA & Qwen2.5 & GPT-4o & Avg & LLaVA & Qwen2.5 & GPT-4o & Avg & LLaVA & Qwen2.5 & GPT-4o & Avg\\ 
    \midrule
    Simulatability  & $0.288$ & $0.219$ & $0.302$ & $0.270$ & $0.305$ & $0.292$ & $0.494$ & $0.364$ & $0.446$ & $0.418$ & $0.357$ & $0.407$ & $0.347$\\
    Informativeness & $0.332$ & $0.152$ & $0.228$ & $0.237$ & $0.429$ & $0.168$ & $0.446$ & $0.348$ & $0.216$ & $0.706$ & $0.538$ & $0.487$ & $0.357$\\
    Plausibility  & $0.162$ & $0.270$ & $0.317$ & $0.250$ & $0.110$ & $0.310$ & $0.325$ & $0.248$ & $0.318$ & $0.391$ & $0.272$ & $0.327$ & $0.275$\\
    \midrule
    Visual Fidelity & $0.197$ & $0.137$ & $0.099$ & $\underline{0.144}$ & $0.271$ & $0.139$ & $0.160$ & $\underline{0.190}$ & $0.558$ & $0.535$ & $0.412$ & $0.501$ & $0.279$\\
    Contrastiveness & $0.176$ & $0.136$ & $0.210$ & $0.174$ & $0.294$ & $0.292$ & $0.424$ & $0.337$ & $0.221$ & $0.276$ & $0.256$ & $0.251$ & $0.254$\\
    \midrule
    Avg(VF, Contr.) & $0.109$ & $0.054$ & $0.090$ & $\textbf{0.084}$ & $0.139$ & $0.138$ & $0.216$ & $\textbf{0.164}$ & $0.360$ & $0.389$ & $0.279$ & $0.343$ & $\textbf{0.197}$\\
    Prod(VF, Contr.) & $0.164$ & $0.154$ & $0.233$ & $0.184$ & $0.290$ & $0.296$ & $0.426$ & $0.337$ & $0.144$ & $0.203$ & $0.200$ & $\textbf{0.182}$ & $0.234$ \\
    Min(VF, Contr.) & $0.147$ & $0.144$ & $0.227$ & $0.173$ & $0.265$ & $0.284$ & $0.425$ & $0.325$ & $0.154$ & $0.234$ & $0.215$ & $\underline{0.201}$ & $\underline{0.233}$ \\
    \bottomrule
\end{tabular}
}
\caption{Expected Calibration Error (ECE, lower is better) for different explanation quality scoring functions across three models and three visual reasoning tasks.}
\label{tab:ece}
\end{table*}

\paragraph{Models.} 
We experiment with three popular vision-language models: \llava \cite{liu2023visual}, \qwen \cite{bai2025qwen2}, and \gpt~\cite{hurst2024gpt}\footnote{We use the \texttt{gpt-4o-2024-05-13} checkpoint.}. 
We mainly use a two-step generation process: models first predict an answer, then generate an explanation for its prediction.
\Cref{sec:apx_ans_prompts} contains the exact prompts we use for answer and explanation generation.

Additionally, our quality scoring functions use several model-based tools. 
For computing Visual Fidelity, \gpt\ is used both for generating verification questions and answering those questions. 
For computing Contrastiveness, we use \gpt\ to mask answers in the explanation and paraphrase question-answer pairs into declarative sentences, and the entailment model from~\citet{sanyal2024machines}.

\paragraph{Baseline Explanation Qualities.} 
We compare our proposed qualities against three established text-only explanation baselines: Simulatability~\cite{hase2020leakage}, Informativeness~\cite{chen2023rev} and Commonsense Plausibility~\cite{liu2023vera}. See \Cref{sec:apx_text_only_metrics_implementations} for details.

\paragraph{Evaluating Quality Score Calibration.} 
We evaluate each quality scoring function on its ability to indicate whether a VLM prediction is accurate based on its explanation. 
Specifically, we consider two evaluation metrics: \textbf{Discriminability (\disc)} and \textbf{Expected Calibration Error (ECE)}.

On an evaluation set of $N$ visual reasoning instances, $\disc(S_\quality)$ evaluates a quality scoring function $S_\quality$ by calculating the difference between the mean quality score assigned to instances with accurate predictions ($\acc(\answer_i) = 1$) and inaccurate predictions ($\acc(\answer_i) = 0$):
\begin{align*}
\disc(S_q)
&= \E_{\substack{1\le i\le N\\ \acc(a_i)=1}}
    \bigl[S_q(E_i)\bigr]\\
&\quad - \E_{\substack{1\le i\le N\\ \acc(a_i)=0}}
    \bigl[S_q(E_i)\bigr].
\end{align*}

We further calculate whether the difference in means between these two distributions ($S_\quality(\rationale)$, for accurate predictions, versus $S_\quality(\rationale)$, for inaccurate predictions) is significant using an unpaired t-test.

Expected Calibration Error (ECE)~\cite{guo2017calibration} is typically used to evaluate how accurately a model's confidence estimate reflects the true accuracy of its predictions. 
We evaluate our quality scoring functions using ECE by interpreting the quality scoring functions as confidence estimates.

\paragraph{Results.} 
\Cref{tab:disc} and \Cref{tab:ece} compare three sets of explanation qualities: existing ones developed for text-only explanations (Simulatability, Informativeness, Plausibility), our proposed qualities (Visual Fidelity, Contrastiveness), and combinations of proposed qualities.
\begin{figure}[t]
    \centering
    \includegraphics[width=\linewidth]{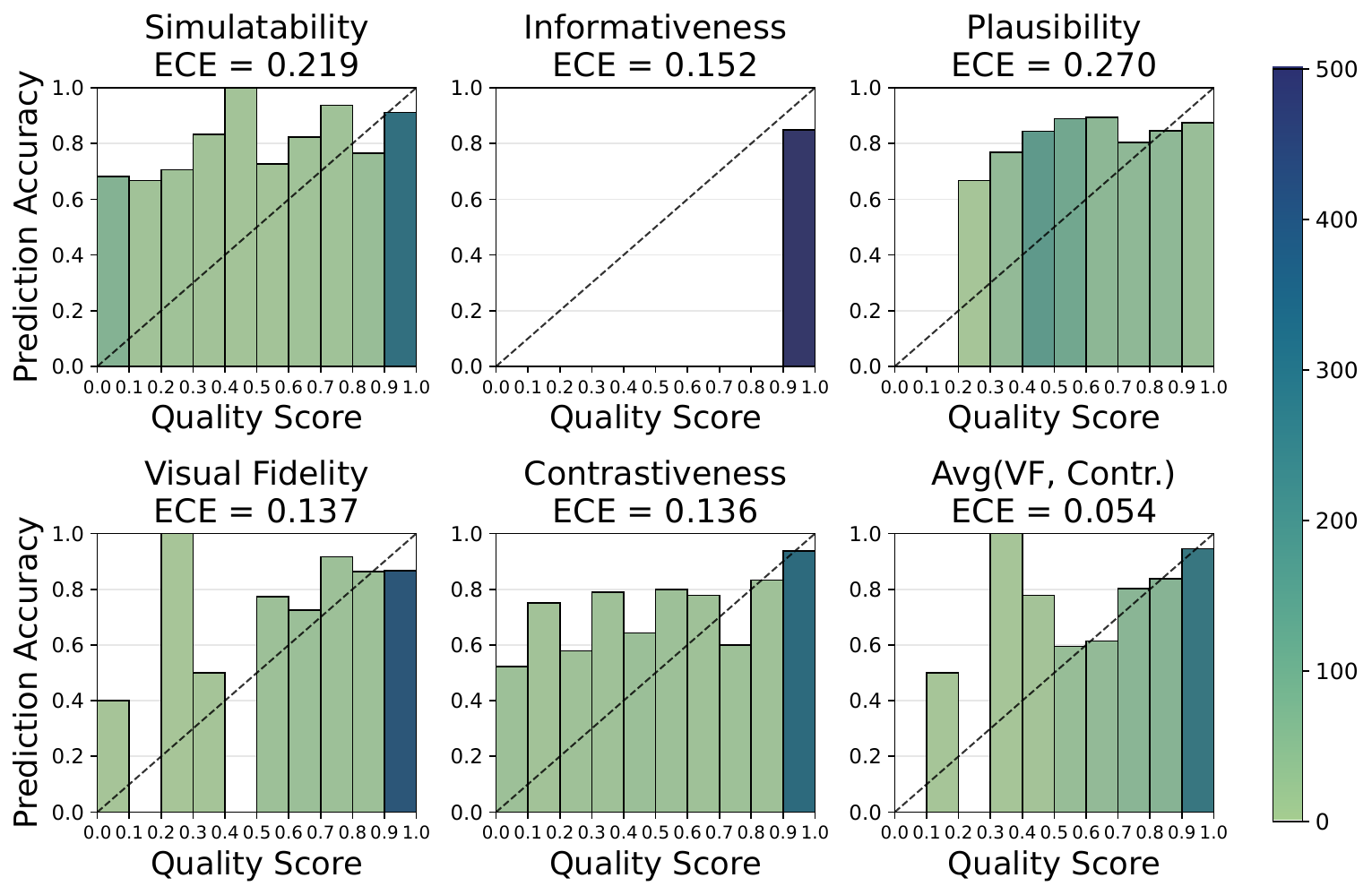}
    \vspace{-20pt}
    \caption{Calibration curves for various quality scoring functions when evaluating explanations generated by \qwen on the A-OKVQA dataset.}
    \vspace{-12pt}
    \label{fig:calib-curves}
\end{figure}
We observe that across all datasets and models, our proposed quality scoring functions almost always achieve higher Discriminability scores and lower Expected Calibration Errors than existing scoring functions across all models. 
Most differences in Discriminability on these datasets are statistically significant ($p < 0.001$), confirming that our metrics effectively distinguish accurate from inaccurate predictions.
We also find that our explanation-based metrics significantly outperform standard logit-based uncertainty baselines on complex reasoning tasks (see \Cref{sec:apx_uncertainty_baselines}).
On VizWiz, while \llava shows strong discriminability, \gpt and \qwen exhibit lower or negative scores on all scoring functions; as discussed in \Cref{sec:apx_vizwiz_negatives}, this is primarily due to valid answers being penalized by the dataset's strict ground truth.
Crucially, \textbf{combining the Visual Fidelity and Contrastiveness quality scores results in the best overall Discriminability and Calibration}, highlighting their complementary utility.

Figure~\ref{fig:calib-curves} shows calibration curves for the baseline qualities, our proposed qualities, and an average of Visual Fidelity and Contrastiveness, for explanations generated by \qwen on the A-OKVQA task. 
We observe that Informativeness's low calibration error is achieved by assigning high scores to all explanations, while our proposed quality scores achieve a lower calibration error while assigning a wider spread of scores.

\paragraph{Verifier robustness.}
To test whether VF's behavior depends on the specific verifier model, we replace the default GPT-4o with two smaller open verifiers (Gemma-3n-E4B and Qwen2.5-VL-7B). 
\Cref{sec:verifier_robustness} shows that all three yield comparable discriminability and calibration, indicating VF remains stable across verifier architectures. 


\section{Are VLM Explanation Quality Scores Helpful to Users?}
\label{sec:user_studies}
Calibration alone does not necessarily translate to downstream utility for users in real-world scenarios~\cite{vodrahalli2022uncalibrated}.
We now investigate whether providing $\visualfidelity$ and $\contrastiveness$ scores alongside VLM explanations helps users more accurately decide when to believe VLM predictions.

\subsection{User Study Setup}
\label{sec:user_study_setup}
We design a user study simulating a real-world setting of a user relying on a VLM assistant where the user cannot directly view the VLM's visual input. 
Study participants are tasked with assessing the accuracy of VLM predictions.

\begin{figure}[t]
    \includegraphics[width=\linewidth]{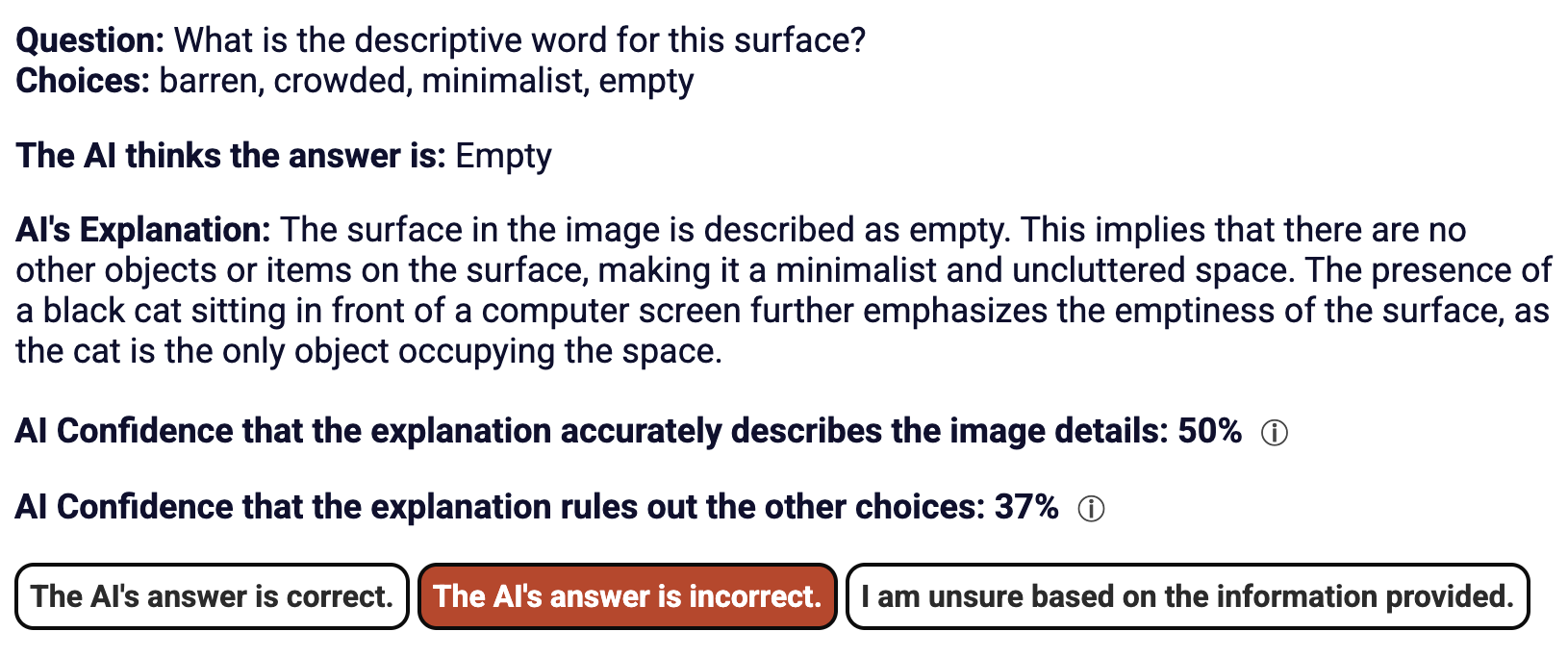}
    \vspace{-23pt}
    \caption{Our study interface where users are shown the visual reasoning question, VLM prediction and explanation, and optionally one or more quality scores
    (using simplified language descriptions). Here, based on the explanation's quality scores, the user correctly believes that the VLM prediction is incorrect.
    }
    \vspace{-10pt}
    \label{fig:interface}
\end{figure}
For a visual reasoning question $(\image, \question)$ from the A-OKVQA and VizWiz tasks, participants are shown the question $\question$, the VLM's prediction $\answer$ and explanation $\rationale$, and optionally an explanation quality score $S_\quality(\rationale)$. 
Importantly, participants \emph{cannot see the image $\image$ accompanying the question}. 
Participants have to decide whether they believe the model's answer is correct or incorrect based on the provided context. They can also indicate ``unsure'' if they feel the information is insufficient for making a reliable judgment. 
This design helps us measure how presenting different quality scores to a user affects user decision making.
Figure~\ref{fig:interface} contains an example of how information is presented in our user study interface.

\paragraph{Study Details.} 
We conducted a between-subjects study on a balanced subset of 100 questions each from A-OKVQA and VizWiz, with 30 participants and 300 annotations per dataset per condition. Participants were recruited via Prolific and paid \$2 base + performance bonuses (see \Cref{sec:apx_user_studies} for full protocol and demographics).

\paragraph{Evaluation Metrics.}
Since participants are asked to judge whether the VLM's prediction is correct or incorrect, we first exclude responses marked as ``unsure.'' 
Following prior work on evaluating human-AI collaboration~\cite{joshi-etal-2023-machine,ma2024arereallysureunderstanding,srinivasan2025adjust,wiegreffe-etal-2022-reframing}, we then calculate \textbf{User Accuracy} over the remaining responses, and the degree of \emph{appropriate reliance} using two metrics: Over-Reliance and Under-Reliance.
\textbf{Over-Reliance} represents the fraction of interactions where the user believed the VLM prediction was correct, when in fact it was incorrect. 
Similarly, \textbf{Under-Reliance} is the fraction of interactions where the user mistakenly believed the VLM was incorrect. 
For both reliance metrics, lower values are preferred.

\subsection{RQ1: Does providing explanation quality scores improve user reliance?}
\label{subsec:userstudy_experiments}
We first evaluate whether showing users our proposed quality scores alongside VLM explanations helps users more accurately assess VLM correctness.
We compare communicating: 1) only the Visual Fidelity score, 2) only the Contrastiveness score, 3) both scores, and 4) a product of the two scores. 
\autoref{fig:all_setting_examples} contains examples of how these quality scores are communicated to users. 
We additionally include a \emph{control} setting where only an explanation is shown without a quality score, and a \emph{random} setting where users are shown a quality score drawn uniformly at random from $[0, 1]$, presented with the same UI as the ``simple confidence'' variant in \Cref{sf:single_num}.

\begin{figure*}[!t]
    \centering
    \includegraphics[width=\linewidth]{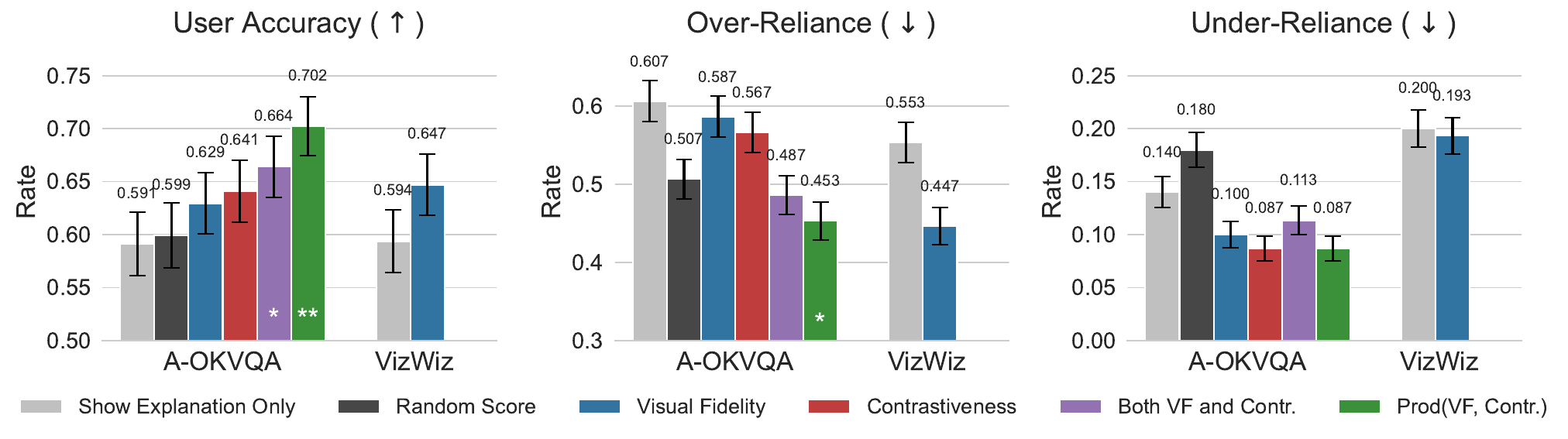}
    \vspace{-24pt}
    \caption{Effect of showing users different quality scores on User Accuracy, Over-Reliance and Under-Reliance. Error bars represent the standard deviation of the data. Asterisks denote improvements over the explanation‐only baseline using a bootstrap significance test (*: $p<0.05$, **: $p<0.01$).
    }
    \vspace{-10pt}
    \label{fig:acc_reliance}
\end{figure*}

\paragraph{Results.}
Presenting explanation quality scores, either individually or in combination, consistently improves user performance across all three metrics (Figure~\ref{fig:acc_reliance}).
Relative to the control explanation-only condition and the Random Score baseline, all treatment conditions improve user accuracy. 
On A-OKVQA, the most pronounced improvements are observed when showing the product of Visual Fidelity and Contrastiveness, which leads to a 11.1\% increase in user accuracy, a 15.4\% reduction in over-reliance, and a 5.3\% reduction in under-reliance. 
These gains suggest that quality signals help users better identify correct model predictions and avoid being misled by incorrect ones.

Other treatment variants also show meaningful improvements.
Visual Fidelity and Contrastiveness alone both reduce over-reliance by approximately 2--4\%, and displaying both scores side-by-side further improves user outcomes.
On VizWiz, the trends are similar; addition of quality scores leads to improvements in both accuracy and over-reliance.
Together, \textbf{these findings highlight the utility of explanation quality scores in helping users make more informed and calibrated decisions about when to trust VLM predictions}.\looseness=-1

\begin{figure}[t]
    \centering
    \includegraphics[width=\linewidth]{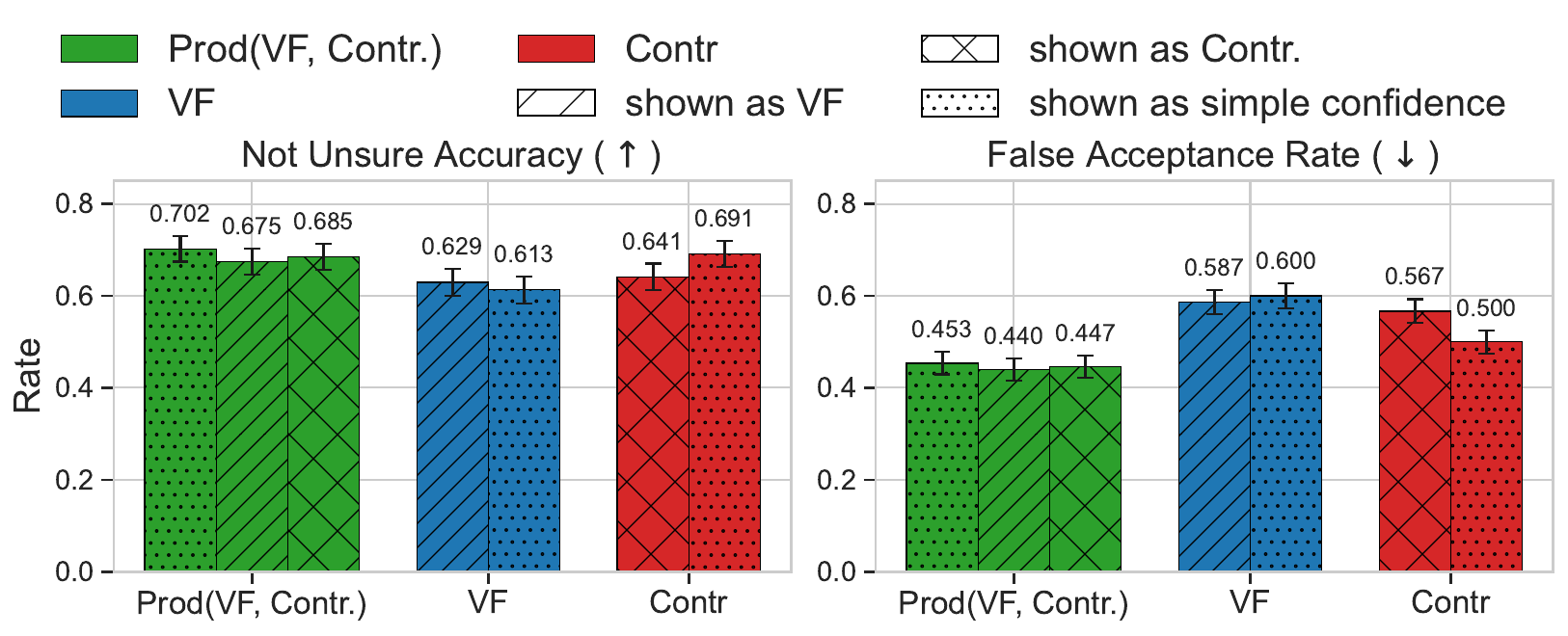}
    \vspace{-22pt}
    \caption{Ablation on \emph{presentation types} on A-OKVQA dataset: holding the numeric signal fixed (Prod($\visualfidelity$, $\contrastiveness$), $\visualfidelity$, or $\contrastiveness$), we vary the on-screen label (\emph{shown as VF}, \emph{shown as Contr}, or \emph{simple confidence}). 
    }
    \vspace{-10pt}
    \label{fig:presentation_types}
\end{figure}

\subsection{RQ2: Which aspects of the presentation affect user decision accuracy?}
\label{subsec:presentation_overview}
Building on our findings in RQ1, we explore which \emph{presentation choices} are significant. We explore two dimensions:
\begin{enumerate}[label=(\roman*).]
    \item Score labeling (RQ2a): We examine whether presenting the same numeric value, but varying its labeling on the study screen (e.g., presenting the score as $\visualfidelity$, $\contrastiveness$, or a generic “AI confidence”) affects user reliance.
    \item Score descriptiveness (RQ2b): Here, we examine the effect of presenting a scalar quality score versus a descriptive summary that presents evidence (e.g., verified visual claims for $\visualfidelity$ or plausible alternatives for $\contrastiveness$).
\end{enumerate}

\subsubsection{RQ2a: Does framing of the quality score affect user reliance?}
\label{subsec:presentation_types}
One concern is whether our observed gains stem from \emph{what} the score communicates versus \emph{how} it is labeled to users. 
For instance, presenting a quality score as the visual fidelity of the explanation may yield different user reliance behavior than if the same score is presented as a generic AI confidence in the explanation.

In this ablation study, using the same underlying score distributions for A-OKVQA predictions (Prod($\visualfidelity$, $\contrastiveness$), $\visualfidelity$, and $\contrastiveness$), we permuted their on-screen presentation: (i) shown under the correct name, (ii) mislabeled as the other qualities or shown as a generic ``AI confidence'' that does not specifically mention either explanation quality. 

\paragraph{Results.}
As shown in Figure~\ref{fig:presentation_types}, renaming/genericizing the numeric signal yields nearly identical outcomes in both User Accuracy and Over-Reliance: differences are small and largely within error bars across all three bases. 
Thus, we observed that user behavior is driven by the numeric signal's distribution rather than its label.

\begin{figure}[t]
    \centering
    \includegraphics[width=\linewidth]{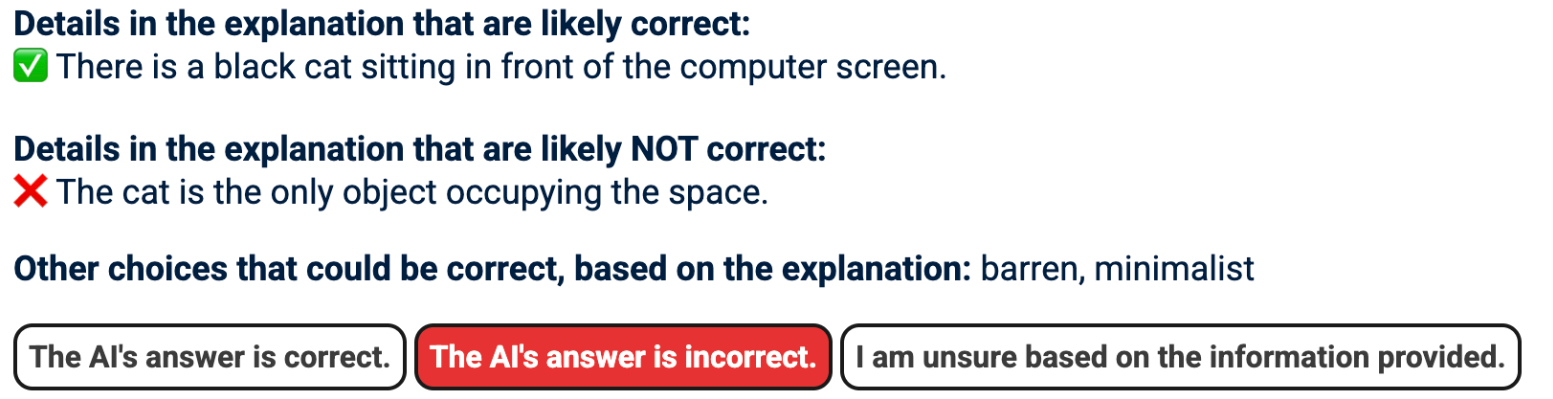}
    \vspace{-20pt}
    \caption{The descriptive quality presentation. The question and explanation display are identical to \Cref{fig:interface}, but numeric scores are replaced with text descriptions of the Visual Fidelity and Contrastiveness qualities.}
    \label{fig:interface_desc}
\end{figure}

\begin{figure}[t]
    \centering
    \vspace{-8pt}
    \includegraphics[width=\linewidth]{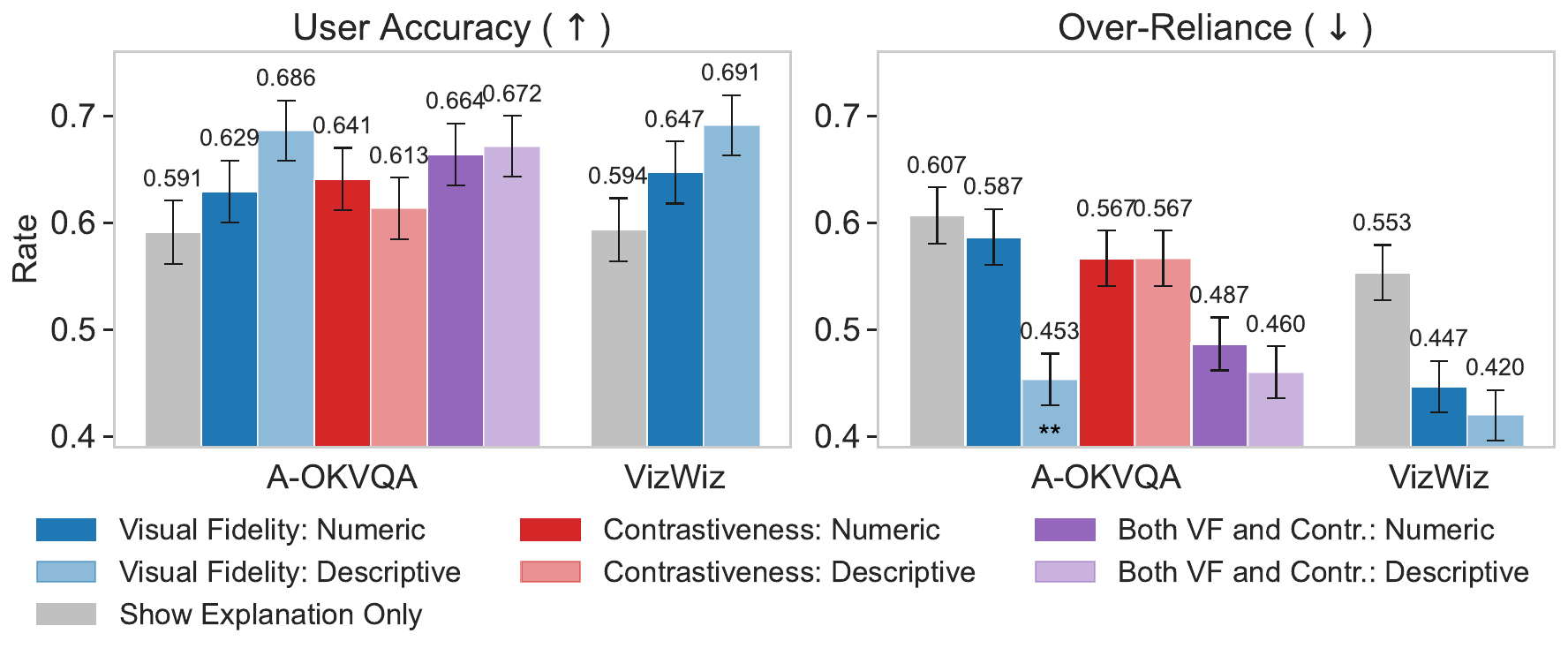}
    \vspace{-20pt}
    \caption{Effect of showing numeric and descriptive qualities on User Accuracy and Over-Reliance. Error bars show standard deviation of the data. Asterisks denote improvements between descriptive setting against the corresponding numeric setting with a bootstrap significance test (**: $p<0.01$).}
    \vspace{-10pt}
    \label{fig:acc_rel_num_dep}
\end{figure}

\subsubsection{RQ2b: Does providing language descriptions of quality scores improve user decision accuracy?}

Our proposed quality scores are interpretable. 
For Visual Fidelity, we can observe which visual details in the explanation were and were not verified by the verifier VLM.
For Contrastiveness, we can examine which alternative answers were also assigned a high entailment probability. 
We compare the effect of showing explanation qualities as numeric scores versus interpretable text descriptions on user decision making.
Figure~\ref{fig:interface_desc} shows an example of descriptive presentations of our proposed qualities.

\paragraph{Results.}
As shown in Figure~\ref{fig:acc_rel_num_dep}, we observe that descriptive formats perform comparably to, and in some cases slightly better than, their numeric counterparts.
While showing both quality scores and for $\visualfidelity$ individually, the descriptive versions lead to improved user accuracy and reduced over-reliance.
However, descriptive treatments hinder performance for $\contrastiveness$, leading to reduced accuracy.
These findings suggest that while numeric scores offer a compact signal, \textbf{descriptive formats may help reduce over-reliance by making the explanation evaluation more transparent}. 

\subsection{RQ3: How does calibration of quality scores impact user decision making?}
Our results in Figure~\ref{fig:acc_reliance} compared quality scores with different levels of calibration, which were also presented with different messages to users. 
To isolate the relationship between calibration error and user decision accuracy from the presentation of explanation qualities, we run user studies by showing scores from different distributions (Visual Fidelity scores, Contrastiveness scores, and average and product combinations of the two) while presenting the scores to users using the same message (\emph{``AI Confidence that the explanation is accurate''}). 

\begin{figure}
    \centering
    \includegraphics[width=\linewidth]{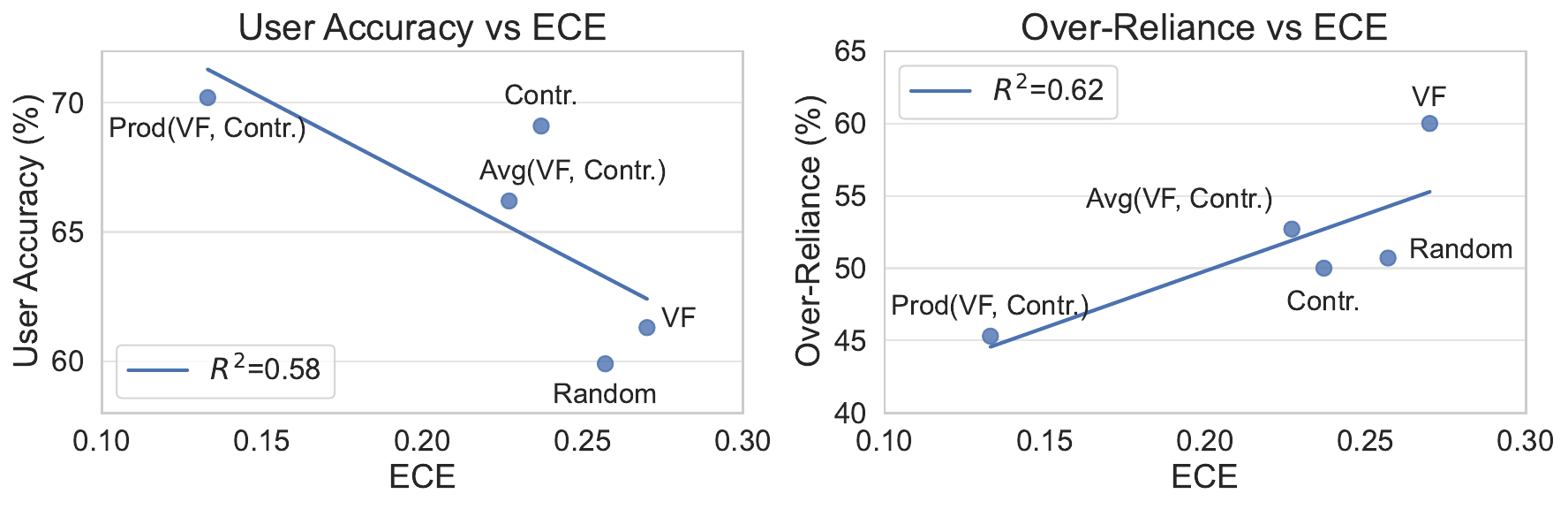}    
    \vspace{-22pt}
    \caption{Relationship between ECE of different quality scores and their downstream utility to users.}
    \vspace{-12pt}
    
    \label{fig:acc_overrel_vs_ECE}
\end{figure}

\paragraph{Results.} 
In Figure~\ref{fig:acc_overrel_vs_ECE}, we compare the ECE of different explanation qualities\footnote{ECE is computed over the 100 samples used in the user study, not the full evaluation set of 500 samples.} and the resulting User Accuracy and Over-Reliance. 
We observe that there is a negative correlation between calibration error and the downstream user decision making accuracy, with ECE explaining $\approx 60\%$ of the variance in User Accuracy and Over-Reliance respectively. 
These findings indicate that \textbf{calibration error of the quality scores is an important determiner of downstream utility to users}.

\section{Conclusions and Future Directions}
\label{sec:conclusions}
While explanations are thought to help users discern when to rely on VLM predictions, providing explanations alone can mislead users into believing the model even when it is incorrect. 
We propose evaluating two unexplored, complementary explanation qualities: Visual Fidelity and Contrastiveness, and also introduce scoring functions for measuring both qualities. 
We find that our proposed quality scoring functions are well calibrated with model correctness, compared to existing notions of explanation quality.
Through several user studies, we demonstrate real-world utility of these scores, as presenting users with quality scores alongside explanations improves their task accuracy and reduces over-reliance on the VLM. 

An important direction is to develop adaptive human-AI reliance strategies that learn when to present explanations and quality scores and when to suppress them, potentially utilizing bandit algorithms to automate system-side abstention~\cite{feng2022learning}. 
Additionally, explanation quality scores could be used as training objectives to improve the generation of explanations. 
Finally, future works should study how explanation quality influences user trust over time.

\section*{Limitations}
\label{sec:limitations_future_work}
One limitation of our approach lies in the open-ended Contrastiveness evaluation. Its score quality is inherently partially bounded by the semantic quality of the generated hard negatives. While we validate that our generator produces challenging distractors in \Cref{sec:apx_vizwiz_negatives}, the metric remains sensitive to their plausibility; if distractors are too easily distinguished, the metric may overestimate confidence, whereas if they are semantically ambiguous, it may penalize valid reasoning. Thus, our metric's resolution is coupled with the generator's capability.
Second, while we demonstrate effectiveness on A-OKVQA, its multiple-choice format may not strictly be representative of real-world, open-ended visual queries. We attempt to bridge this by evaluating on open-ended VizWiz examples and the diverse MMMU-Pro benchmark, but future work should explore even broader task distributions.
Finally, we only evaluate on English-language datasets and conduct user studies with fluent English speakers. Future work should explore multilingual generalization.

\section*{Ethics Statement}
\label{sec:ethics_statement}
This work was conducted with careful consideration of ethical implications. 
Human studies were conducted with informed consent through Prolific, with 
participants compensated at rates exceeding US minimum wage (\$2 base + performance bonuses, see \Cref{subsubsec:apx_bonus_payments}). 
Participants were native English speakers from the US with higher education credentials (\Cref{tab:annotator_filtering}). 
We acknowledge this may limit generalizability across demographic groups.

While our work aims to improve accessibility for blind and low-vision users by calibrating trust in VLM outputs, we recognize potential risks: 
(1) users might over-rely on quality scores themselves, 
(2) errors in quality assessment could disproportionately affect vulnerable populations, and 
(3) current English-only implementation excludes non-English speakers. 
Future work should address these limitations through multilingual support and extensive testing with blind/low-vision communities.

The VizWiz dataset used contains questions from blind and low-vision users but is fully anonymized. In our human studies, the annotators were anonymized. We filtered NSFW content (\Cref{subsec:create_subset_split}) and did not collect any additional personal information. All model outputs were manually reviewed for offensive content before presentation to study participants.

A-OKVQA and MMMU-Pro are released under Apache-2.0 license. VizWiz is released under a CC BY 4.0 license. \gpt is accessed via OPENAI API under their terms of service.
\llava and \qwen are both under Apache-2.0 license. \hyperlink{https://huggingface.co/soumyasanyal/nli-entailment-verifier-xxl}{soumyasanyal/nli-entailment-verifier-xxl} model we used for the NLI task (see \Cref{subsec:apx_support}) is under a CC BY 4.0 license.

Our use of A-OKVQA, VizWiz, and MMMU-Pro aligns with their intended research purposes for evaluating VQA systems. We use language models consistent with their intended purposes.

We used AI assistants (e.g., ChatGPT) for grammar checking and code debugging, but not for producing substantive scientific content or analysis. All model outputs were reviewed and verified by the authors.

\section*{Acknowledgments}
This research is supported in part by the National Science
Foundation (NSF) under grant IIS2403437, the Simons Foundation, the Allen Institute for AI, DARPA award HR00112490376, and a Capital One CREDIF research grant. The views and conclusions contained herein are those of the authors and
should not be interpreted as necessarily representing the official policies, either expressed or implied,
of NSF.
B. Joshi was supported by the Apple Scholars in
AI/ML PhD Fellowship. This work was partially
done when S. Swayamdipta was a visitor at the
Simons Institute for the Theory of Computing.

\bibliography{custom}

\newpage
\appendix

\section{Related Work}
Our work builds on insights from evaluation of explanation quality and their utility for decision support, and adapt them for a vision-language setting.

\paragraph{Evaluating Explanation Quality.}
Prior work on evaluating explanation quality has largely focused on text-only domains, proposing metrics that assess informativeness~\cite{chen2023rev}, prediction support~\cite{wiegreffe2022measuringassociationlabelsfreetext, hase2020leakage}, acceptability or helpfulness to users~\cite{wiegreffe-etal-2022-reframing}, and consistency across reasoning chains~\cite{golovneva2023roscoesuitemetricsscoring, prasad2023recevalevaluatingreasoningchains}. 
Our work goes beyond textual plausibility by introducing two novel explanation qualities tailored for vision-language settings. 
Visual Fidelity has been explored for evaluating faithfulness of model-generated image captions~\cite{madhyastha2019vifidelevaluatingvisualfidelity,lee2023aligningtexttoimagemodelsusing}. 
Our work builds on these ideas to do zero-shot evaluation of explanations.
Separately, recent work has investigated the calibration of verbalized confidence from models directly~\cite{xuan2025seeingbelievingmuchcomprehensive}, whereas we derive calibration signals from the explanation content itself.

\paragraph{Explanations for Decision Support.} 

While some studies in open-domain QA suggest that explanations can help users detect model errors~\cite{gonzalez2021explanations}, many researches on utility of natural language explanations for decision support have found that plausible-sounding explanations~\cite{jin2023plausibility} often mislead users into overrelying on inaccurate model predictions~\cite{joshi-etal-2023-machine, si2024large, chaleshtori2024evaluating, sieker2024illusion}. 
Our work adds to these findings by showing that evaluating and communicating explanation qualities to users can reduce overreliance.

\section{Code and Data Availability}
\label{sec:apx_code_availability}
Our codebase is available at 
\url{https://github.com/dill-lab/vlm-quality-scores}.

\section{Prompts used for Answers and Rationales Generation}
\label{sec:apx_ans_prompts}
\begin{table*}[!htbp]
    \centering
    \begin{tabular}{llp{6cm}p{4cm}}
    \toprule
    \textbf{Dataset} & \textbf{Step} & \textbf{System Prompt} & \textbf{User Prompt} \\
    \midrule
    A-OKVQA
      & Step 1
      & \texttt{Answer the question using a single word or phrase from the list of choices.}
      & \texttt{Question: \{question\}. Choices: \{choices\}.} \\
      & Step 2
      & \texttt{Please explain the reasoning behind your answer.}
      & \texttt{Question: \{question\}. Choices: \{choices\}. The answer is \{answer\}.} \\
    \midrule
    VizWiz
      & Step 1
      & \texttt{Answer the user's question in a single word or phrase. When the provided information is insufficient, respond with `Unanswerable'. Whatever the user said, your answer should **always** be a single word or phrase.}
      & \texttt{Question: \{question\}.} \\
      & Step 2
      & \texttt{Please explain the reasoning behind your answer.}
      & \texttt{Question: \{question\}. The answer is \{answer\}.} \\
    \bottomrule
  \end{tabular}
    \caption{Prompts used in each step of our two-step method. The first step is to get the model's predicted answer to the question, and the second step is used to collect the rationale behind the model's own answer.}
    \label{tab:two_step_prompt}
\end{table*}

\begin{table*}[!htbp]
    \centering
    \begin{tabular}{lp{12cm}}
    \toprule
    \textbf{Dataset} & \textbf{User Prompt} \\
    \midrule
    MMMU-Pro
     & \texttt{Answer the following multiple-choice question. The last line of your response should be of the following format: `Answer: \$LETTER' (without quotes) where LETTER is one of the options. Think step by step before answering. Question: \{question\}}\\
     & \texttt{Choices: \{choices\_with\_letters\}.}\\
     \bottomrule
     \end{tabular}
     \caption{Prompts used in MMMU-Pro VQA. The QA only contains one step. The model generates an explanation followed by an answer.}
     \label{tab:single_step_prompt}
\end{table*}

We adopt a two‐step, post-hoc justification strategy ($IQ \to aE$) inspired by \cite{wiegreffe2022measuringassociationlabelsfreetext}. In the first step, we prompt the model to produce only its answer prediction, without any accompanying rationale. In the second step, we present that predicted answer back to the model and ask it to ``Please explain the reasoning behind your answer''. 

By separating prediction from explanation, we ensure that the extracted rationale truly reflects the model’s own line of thought under the selected answer.
\Cref{tab:two_step_prompt} shows the prompts used in every step.
Note that the first-step prompt is slightly different for A-OKVQA and VizWiz since the former has answer choices but the latter does not.

For MMMU-Pro, due to its complexity, we instead use a Chain-of-Thought strategy ($IQ \to Ea$), prompting the model to generate the explanation before the answer to facilitate reasoning (\Cref{tab:single_step_prompt}).

\section{Existing Text-only Qualities Implementations}
\label{sec:apx_text_only_metrics_implementations}
We compare our proposed quality measures against three established, text-only explanation qualities, each of which also returns a score in the range $[0, 1]$. 
\textbf{Simulatability}~\cite{hase2020leakage} evaluates whether an explanation offers sufficient evidence to logically justify the model's prediction. 
\textbf{Informativeness}~\cite{chen2023rev} evaluates whether the explanation introduces new information to justify a prediction beyond just re-stating the question and prediction. 
\textbf{Commonsense Plausibility}~\cite{liu2023vera} evaluates whether the explanation is in accordance with commonsense knowledge of everyday situations. 

\subsection{Simulatability}
\label{subsec:apx_support}
To mitigate answer leakage in our Simulatability quality, we first mask all direct mentions of the model’s predicted answer within each rationale by replacing them with the special token \texttt{<mask>}.
Given the model's predicted answer $a$ and its rationale $E$, every occurrence of $a$ in $E$ is substituted with \texttt{<mask>}, yielding the masked rationale $E_{masked}$.

Next, to transform the QA pair into an NLI task, we prompt \gptmini to convert the original question and its predicted answer into a single declarative hypothesis sentence $H$. \Cref{tab:desc_sentence_gen} shows the model and prompt used to generate $H$. We then feed the masked rationale $E_{masked}$ as the premise and $H$ as the hypothesis into the \hyperlink{https://huggingface.co/soumyasanyal/nli-entailment-verifier-xxl}{soumyasanyal/nli-entailment-verifier-xxl} model on Hugging Face to obtain an entailment probability $p_{entail}$ as its Simulatability score.

\begin{table*}[t]
  \centering
  \begin{tabular}{cp{13cm}}
    \toprule
    \textbf{Config} & \textbf{Assignment} \\
    \midrule
    Model & \texttt{gpt-4o-mini-2024-07-18} \\
    \midrule
    Max Tokens & 1024 \\
    Temperature & 0.1 \\
    \midrule
    \midrule  
    User Prompt &
      \begin{minipage}[t]{\linewidth}
\texttt{Integrate the question and the answer into one sentence.
For example, given the question "What is the man waiting for?" and the answer "taxi", you should output "The man is waiting for taxi."
\\
Question: \{question\}\\
Answer: \{answer\}}
      \end{minipage} \\
    \bottomrule
  \end{tabular}
    \caption{Model configuration and prompt used to generate descriptive sentence}
  \label{tab:desc_sentence_gen}
\end{table*}

\subsection{Informativeness}
We utilized \gpt to extract the new information contained in the model’s rationale that are not semantically equivalent to the hypothesis.  
\begin{table*}[t]
  \centering
  \begin{tabular}{cp{12.5cm}}
    \toprule
    \textbf{Config} & \textbf{Assignment} \\
    \midrule
    Model & \texttt{gpt-4o-2024-05-13} \\
    \midrule
    Max Tokens & 1024 \\
    Temperature & 0.1 \\
    \midrule
    \midrule  
    User Prompt &
      \begin{minipage}[t]{\linewidth}
\texttt{Please break the following rationale into distinct pieces, and keep only the ones that are not semantically equivalent to the hypothesis. Output the final answer in a Python list format.\\ \newline
Example:\\
Hypothesis: The man by the bags is waiting for a delivery.\\
Rationale: The man by the bags is waiting for a delivery, as indicated by the presence of the suitcases and the fact that he is standing on the side of the road. The other options, such as a skateboarder, train, or cab, do not seem to be relevant to the situation depicted in the image.\\
Output: ["Suitcases are present in the image.", "The man is standing on the side of the road.", "The other options, such as a skateboarder, train, or cab, do not seem to be relevant to the situation depicted in the image."]\\
\newline
Task:\\
Hypothesis: \{hypothesis\}\\
Rationale: \{rationale\}}
      \end{minipage} \\
    \bottomrule
  \end{tabular}
    \caption{Model configuration and prompt used to evaluate informativeness of a rationale.}
  \label{tab:inform_prompt}
\end{table*}

\Cref{tab:inform_prompt} shows the model configuration and prompt used to evaluate informativeness of a rationale. After extracting the individual information pieces, we check the size of the resulting list; if it is non–zero, we deem the rationale to be informative.

\subsection{Commonsense Plausibility}
To evaluate the plausibility of the model's explanations against commonsense knowledge, we utilize the Vera model~\cite{liu2023vera} directly, without modification.

\section{Visual Fidelity}
\label{sec:apx_VF}

In Visual Fidelity, we use a two-step pipeline to evaluate. Firstly, we generate possible visual verification questions related to the rationale, by providing the question, predicted answer, and rationale to the model, and second, for each question, we provide the question and visual input to the model to ask for verification.

\begin{table*}[t]
  \centering
  \small  
  \begin{tabular}{cp{12cm}}
    \toprule
    \textbf{Config} & \textbf{Assignment} \\
    \midrule
    Model & \texttt{gpt-4o-2024-08-06} \\
    \midrule
    Max Tokens & 1024 \\
    Temperature & 0.1 \\
    \midrule
    \midrule  
    \multicolumn{2}{l}{%
      \begin{minipage}[t]{\linewidth}
User prompt:
\texttt{You will be shown a question about an image, along with an answer, and a rationale that explains the answer based on details from the image. Your task is to generate a list of yes/no questions that verify the details about the image that are **explicitly** mentioned in the rationale. Your questions should be phrased such that the answer to that question being yes means that the detail in the rationale is correct. Focus on creating questions that can be visually verified or refuted based on the details provided in the rationale. Ensure the questions are specific and directly pertain to aspects that are visually relevant and mentioned in the rationale. Avoid generating questions about elements that are not mentioned in the rationale, or the rationale explicitly states are not relevant or present. Also avoid generating multiple questions that check for the same visual detail.\\
\newline
Here is one example:\\
Input: \\
Question: Why is the person wearing a helmet?\\
Answer: For safety\\
Rationale: The person is wearing a helmet because they are riding a bicycle on a busy city street. Helmets are commonly used to protect against head injuries in case of accidents, especially in areas with heavy traffic.\\
\newline
Good Questions:\\
1. Is the person wearing a helmet while riding a bicycle?\\
Reason: This question is directly answerable by observing whether the person on the bicycle is wearing a helmet in the image. \\
2. Is the street in the image busy with traffic?\\
Reason: This question can be visually verified by looking at the amount of traffic on the street in the image.\\
\newline
Bad Questions:\\
1. Is the person wearing the helmet because they are concerned about head injuries?\\
Reason: This question is not good because it assumes the person’s intentions or concerns, which cannot be visually verified from the image.\\
2. Does wearing a helmet suggest that the person is highly safety-conscious?\\
Reason: This question relies on inference and external knowledge about the person’s mindset, rather than on observable details from the image.\\
3. Is there any indication that the person is wearing a helmet for safety reasons?\\
Reason: This question verifies the answer to the original question, rather than verifying a detail about the image that's mentioned in the rationale.\\
4. Is the person wearing a safety vest?\\
Reason: This question is not good because it tries to verify details about the image that are not explicitly mentioned in the rationale.\\
5. Is the person not wearing sunglasses?\\
Reason: This question is not good because it asks for verification by absence and can only be answered with a "no," which is not the preferred type of question.\\
\newline
Respond with a list of (good) questions (without the reasons), starting from `1. '}
      \end{minipage} %
      }\\
    \bottomrule
    \end{tabular}
      \caption{Model configuration and prompt used to generate verification visual questions of a rationale.}
  \label{tab:VF_prompt_1}
\end{table*}
\begin{table*}[!htbp]
  \centering
  \begin{tabular}{cp{11cm}}
    \toprule
    \textbf{Config} & \textbf{Assignment} \\
    \midrule
    Model & \texttt{gpt-4o-2024-08-06} \\
    \midrule
    Max Tokens & 1024 \\
    Temperature & 0.1 \\
    \midrule
    \midrule  
    \multicolumn{2}{l}{%
      \begin{minipage}[t]{\linewidth}
User prompt:
\texttt{Question: \{question\}. Based on the information provided in the image, answer with `yes' or `no'. Provide one-word answer only.}
      \end{minipage} %
      }\\
    \bottomrule
    \end{tabular}
      \caption{Model configuration and prompt used to verify the visual questions generated from \Cref{tab:VF_prompt_1}.}
  \label{tab:VF_prompt_2}
\end{table*}

\Cref{tab:VF_prompt_1} refers to the model setting and prompt used for generating the verification questions, while
\Cref{tab:VF_prompt_2} refers to those used for verifying the questions.

\section{Hard Negatives Generations}
\label{sec:apx_hard_negatives}

To enable Contrastiveness evaluation on open-ended datasets like VizWiz (which lack fixed answer choices), we dynamically generate ``Hard Negatives'' to serve as plausible distractors. We use \gpt as the generator, providing it with the question, the ground-truth correct answer(s), and the image context. The model is instructed to create distractors that are incorrectly but visually plausible, explicitly avoiding simple synonyms. This aligns with prior work on automated distractor generation~\cite{lu2022goodbetterbesttextual}; however, unlike previous learning-based methods, we employ a training-free approach using an off-the-shelf LLM. \Cref{tab:hard_negatives_prompt} shows the exact prompt used.

\begin{table*}[ht]
\centering
\small
\begin{tabular}{p{\linewidth}}
\toprule
\textbf{Hard Negatives Generation Prompt} \\
\midrule
Given the following question, its correct answer(s), and the accompanying image, generate exactly THREE (3) different "Hard Negative" answers. \\
\\
Context: \\
Question: \{question\} \\
Correct Answer(s): \{valid\_answers\_str\} \\
\\
Your Goal: Generate 3 "Hard Negative" distractors. \\
A "Hard Negative" must be: \\
1. \textbf{Incorrect}: It must NOT be a synonym or a close grammatical variation of the correct answer(s). \\
2. \textbf{Visually Plausible}: Ideally, it refers to something else \textit{visible in the image} or common in this context. \\
3. \textbf{Challenging}: It should be a reasonable guess for a Model that didn't pay perfect attention effectively. \\
\\
Constraints: \\
- \textbf{NO Synonyms}: If the answer is "bike", do NOT generate "bicycle". \\
- \textbf{Format}: Provide exactly THREE (3) comma-separated phrases. No numbering, no extra text. \\
- \textbf{Length}: Short, concise (1-3 words). \\
\bottomrule
\end{tabular}
\caption{System prompt used to generate hard negatives for Contrastiveness evaluation on open-ended datasets (VizWiz).}
\label{tab:hard_negatives_prompt}
\end{table*}

\section{Validation of Hard Negatives Strategy}
\label{sec:apx_hard_negatives_validation}

To validate our strategy of dynamically generating hard negatives for open-set evaluation (as used in VizWiz), we conducted an ablation study on the A-OKVQA dataset. A-OKVQA typically provides 4 ground-truth answer choices (Closed-Set). We stripped these choices and instead generated hard negatives using the same pipeline applied to VizWiz, effectively creating a ``Reconstructed Closed-Set'' version of A-OKVQA.

\begin{table*}[!htbp]
\centering
\small
\begin{tabular}{llccc}
\toprule
\textbf{Model} & \textbf{Metric} & \textbf{Closed-Set} & \textbf{Reconstructed Closed-Set} & \textbf{Delta} \\
 & & (Original) & (Generated Hard Negatives) & \\
\midrule
\multirow{2}{*}{\gpt} & Disc ($\uparrow$) & $^{\phantom{***}}0.248^{***}$ & $^{\phantom{*}}0.111^{*}$ & $-0.137$ \\
 & ECE ($\downarrow$) & $0.210$ & $0.312$ & $+0.102$ \\
\midrule
\multirow{2}{*}{\qwen} & Disc ($\uparrow$) & $^{\phantom{***}}0.283^{***}$ & $^{\phantom{***}}0.158^{***}$ & $-0.125$ \\
 & ECE ($\downarrow$) & $0.136$ & $0.235$ & $+0.099$ \\
\midrule
\multirow{2}{*}{\llava} & Disc ($\uparrow$) & $^{\phantom{***}}0.243^{***}$ & $^{\phantom{***}}0.112^{***}$ & $-0.131$ \\
 & ECE ($\downarrow$) & $0.176$ & $0.233$ & $+0.057$ \\
\bottomrule
\end{tabular}
\caption{Ablation study on A-OKVQA comparing standard Closed-Set evaluation (using ground-truth options) vs. Reconstructed Closed-Set evaluation (using generated hard negatives) for the Contrastiveness quality. The generated negatives significantly increase task difficulty.}
\label{tab:aokvqa_ablation}
\end{table*}

\Cref{tab:aokvqa_ablation} compares the performance of our Contrastiveness quality on the original Closed-Set vs. the Reconstructed Closed-Set. We observe a consistent drop in Discriminability ($\sim0.13$) and a rise in ECE ($\sim0.08$) across all models. This indicates that the generated hard negatives successfully make the task more challenging, likely because the generated distractors are semantically closer to the correct answer than the original multiple-choice options. However, crucially, the discriminability scores remain positive and statistically significant (e.g., $0.158^{***}$ for \qwen), confirming that our metric remains valid and robust even in this harder, constructed setting.

To further understand why the generated negatives increase difficulty, we analyzed the semantic proximity of the distractors to the correct answer using \texttt{all-mpnet-base-v2}. Across the 500-sample set, the mean cosine similarity for Original Negatives was $0.424$, whereas for Generated Negatives it rose to $0.468$ ($\Delta +0.044$). 

\begin{table*}[t]
\centering
\small
\resizebox{\linewidth}{!}{
\begin{tabular}{p{2.5cm} p{3cm} p{1.5cm} p{4cm} p{4cm}}
\toprule
\textbf{Image} & \textbf{Question} & \textbf{Correct} & \textbf{Original Negatives} (Sim) & \textbf{Generated Negatives} (Sim) \\
\midrule
\includegraphics[width=\linewidth, valign=t]{figs/VQA_examples/AOKVQA-111.jpg} & 
What kind of coating has been used? & 
frosting & 
polish, paint, varnish \newline ($S_{\text{sim}} = 0.27$) & 
buttercream, whipped cream, glaze \newline ($S_{\text{sim}} = 0.58$) \\
\midrule
\includegraphics[width=\linewidth, valign=t]{figs/VQA_examples/AOKVQA-308.jpg} & 
In what type of location are they playing with the body board? & 
room & 
beach, park, store \newline ($S_{\text{sim}} = 0.28$) & 
living room, hallway, garage \newline ($S_{\text{sim}} = 0.58$) \\
\bottomrule
\end{tabular}
}
\caption{Comparison of Original (Closed-Set) vs. Generated (Reconstructed Closed-Set) distractors for matching A-OKVQA questions. The generated negatives are often semantically closer to the correct answer (higher similarity $S_{\text{sim}}$), making the discrimination task harder.}
\label{tab:hard_negatives_comparison}
\end{table*}

While A-OKVQA provides a rigorous baseline with plausible distractors, our Reconstructed Closed-Set method produces even more semantically proximal candidates. As shown in \Cref{tab:hard_negatives_comparison}, the generated options (e.g., ``buttercream'' vs ``frosting'') often require finer-grained visual discrimination than original distractors (e.g., ``paint'' vs ``frosting''). This suggests that LLM-generated hard negatives serve as a harder mode for evaluation, explaining the lower discriminability scores while verifying the metric's robustness efficiently.

\section{Uncertainty Quantification Baselines}
\label{sec:apx_uncertainty_baselines}

To situate our explanation quality metrics within broader uncertainty quantification literature, we compare them against two standard uncertainty baselines. Specifically, for the \llava model (where we have access to output probabilities), we evaluate:
\begin{enumerate}[nosep]
  \item Model Confidence (Logits): the softmax probability of the first token of the predicted answer.
  \item $P(\text{True})$: prompting the model ``Is the proposed answer correct?'' and computing a softmax over the logits of affirmative versus negative tokens.
\end{enumerate}

\begin{table*}[ht] 
\centering 
\small
\begin{tabular}{lp{11cm}}
\toprule
\textbf{Method} & \textbf{Prompt} \\
\midrule
$P(\text{True})$ &
\texttt{Question: \{question\}} \newline
\texttt{Choices: \{choices\}} \newline
\texttt{Proposed answer: \{predicted\_answer\}} \newline\newline
\texttt{Is the proposed answer correct? Respond with only "True" or "False".} \\
\bottomrule
\end{tabular}
\caption{Prompt used for the $P(\text{True})$ uncertainty quantification baseline. We extract the logits of affirmative tokens and negative tokens at the first generation step, taking the maximum within each group, and normalize via softmax to obtain $P(\text{True})$.}
\label{tab:uq_prompts}
\end{table*}

The exact prompts are shown in \Cref{tab:uq_prompts}.

\begin{table*}[!ht]
\centering
\begin{tabular}{lrrrr}
\toprule
 & \multicolumn{2}{c}{\textbf{A-OKVQA} (Simpler)} & \multicolumn{2}{c}{\textbf{MMMU-Pro} (Harder)} \\
\cmidrule(lr){2-3} \cmidrule(lr){4-5}
\textbf{Metric} & \textbf{Disc.} $\uparrow$ & \textbf{ECE} $\downarrow$ & \textbf{Disc.} $\uparrow$ & \textbf{ECE} $\downarrow$ \\
\midrule
Model Confidence (Logits) & $0.252^{***}$ & $\mathbf{0.046}$ & $0.055^{\phantom{***}}$ & $0.356$ \\
$P(\text{True})$ & $0.032^{***}$ & $0.253$ & $-0.001^{\phantom{***}}$ & $0.772$ \\
\midrule
VF & $0.181^{***}$ & $0.197$ & $\mathbf{0.169}^{**\phantom{*}}$ & $0.558$ \\
Contr. & $0.243^{***}$ & $0.176$ & $0.047^{\phantom{***}}$ & $0.221$ \\
Avg(VF, Contr.) & $0.212^{***}$ & $0.109$ & $0.108^{**\phantom{*}}$ & $0.360$ \\
Prod(VF, Contr.) & $\mathbf{0.320}^{***}$ & $0.164$ & $0.090^{**\phantom{*}}$ & $\mathbf{0.144}$ \\
\bottomrule
\end{tabular}

\caption{Comparison of our explanation quality metrics against uncertainty quantification baselines on LLaVA-v1.5-7B. We consider these baseline metrics: logit-based confidence (softmax probability of the predicted answer's first token) and $P(\text{True})$ (probability assigned to ``True'' when asked if the prediction is correct). Significance is evaluated using an unpaired t-test: $^{***}$ $p<0.001$, $^{**}$ $p<0.01$,
$^{*}$ $p<0.05$.}
\label{tab:logits_baseline}
\end{table*}

\Cref{tab:logits_baseline} presents the comparison on two contrasting datasets: A-OKVQA (simpler visual reasoning) and MMMU-Pro (complex, multi-discipline reasoning). 

On A-OKVQA, the model's intrinsic ``intuition'' (logits) is highly effective, achieving strong Discriminability ($0.252$) and excellent ECE ($0.046$). P(True) ($0.032$) exhibits substantially weaker discriminability and poor calibration (ECE $> 0.25$). This suggests that for simpler tasks, raw confidence is a sufficient reliability indicator.

On MMMU-Pro, both uncertainty baselines break down. Logits yield poor discriminability ($0.055$, not significant) and high calibration error ($0.356$); P(True) produces especially poor calibration (ECE of $0.772$). In contrast, our explanation quality metrics---particularly the product of Visual Fidelity and Contrastiveness ($S_{\product}$)---remain robust, achieving statistically significant discriminability ($0.090^{**}$) and much better calibration ($0.144$).

These results indicate that logits (which measure \textit{answer} confidence) and our metrics (which measure \textit{explanation} reliability) are distinct signals. While logits suffice for simpler tasks, explanation quality offers a necessary and robust alternative for complex reasoning, particularly when model confidence becomes unreliable or inaccessible.

\section{Qualitative Examples}
\label{sec:apx_qualitative_examples}

\Cref{tab:example-model-explanations} shows our evaluation framework on four representative VQA instances. For each example, we provide the original question (and answer choices where applicable), image, ground truth answer, the model’s selected answer and rationale, and the corresponding scores for Simulatability, Informativeness, Commonsense Plausibility, Visual Fidelity, and Contrastiveness. Note that for VizWiz, the ground-truth answer is defined as the majority vote among the 10 user annotations from the original dataset.

\begin{table*}[ht]
\centering
\resizebox{\linewidth}{!}{
    \begin{tabular}{%
      p{1.3cm} 
      p{2cm} 
      p{1.5cm} 
      p{3cm} 
      p{1.2cm} 
      p{1.2cm} 
      p{6cm} 
      p{0.7cm} 
      p{0.7cm} 
      p{0.7cm} 
      p{0.7cm} 
      p{0.7cm} 
    }
    \toprule
    \textbf{Dataset} &
    \textbf{Question} &
    \textbf{Answer Choices} &
    \textbf{Image} &
    \textbf{Ground Truth} &
    \textbf{Predicted Answer} &
    \textbf{\llava Explanation} &
    \textbf{Sim.} &
    \textbf{Info.} &
    \textbf{Plau.} &
    \textbf{VF} &
    \textbf{Contr.}\\
    \midrule
    A-OKVQA&
    What is the white substance on top of the cupcakes? &
    butter, mayo, ice cream, icing &
    \includegraphics[width=\linewidth,valign=t]{figs/VQA_examples/AOKVQA-3.jpg} &
    
    icing &
    icing &
    The white substance on top of the cupcakes is icing. This can be inferred from the fact that icing is commonly used as a topping for cupcakes to enhance their flavor and appearance. In the image, there are three cupcakes with icing on top, which indicates that they have been decorated with icing as part of their presentation. Additionally, icing is a common ingredient in cupcake recipes, so it is reasonable to assume that the white substance on top of the cupcakes is indeed icing. &
    0.391&
    1.000&
    0.473&
    1.000&
    0.996\\
    
    A-OKVQA&
    What is the descriptive word for this surface?&
    barren, crowded, minimalist, empty&
    \includegraphics[width=\linewidth, valign=t]{figs/VQA_examples/AOKVQA-7.jpg}&
    crowded&
    empty&
    The surface in the image is described as empty. This implies that there are no other objects or items on the surface, making it a minimalist and uncluttered space. The presence of a black cat sitting in front of a computer screen further emphasizes the emptiness of the surface, as the cat is the only object occupying the space.&
    0.964&
    1.000&
    0.566&
    0.500&
    0.393\\

    VizWiz&
    What color are those pants?&
    \textcolor{gray}{blue, red, green}&
    \includegraphics[width=\linewidth, valign=t]{figs/VQA_examples/Vizwiz-1006.jpg}&
    pink, pink white checkered&
    pink&
    The man is wearing pink pajamas, which is an unusual choice for a casual setting.&
    0.025&
    1.000&
    0.397&
    1.000&
    \textcolor{gray}{0.073}\\

    VizWiz&
    What is in this bottle?&
    \textcolor{gray}{pen, laser pointer, small torch}&
    \includegraphics[width=\linewidth, valign=t]{figs/VQA_examples/Vizwiz-686.jpg}&
    bacon bits, hormel bacon bits, bacon&
    spices&
    The person holding the bottle of spices is likely using it to season their food, adding flavor and aroma to various dishes. Spices are commonly used in cooking as they can enhance the taste of ingredients and create unique flavor combinations. The hand holding the bottle suggests that the person might be preparing a meal or experimenting with different spice blends for their culinary creations.&
    0.999&
    1.000&
    0.566&
    0.500&
    \textcolor{gray}{0.555}\\
    
    \bottomrule
    \end{tabular}
}
\caption{Four VQA examples from both A-OKVQA and VizWiz datasets with model predictions, explanations, and qualities. The predicted answer and explanation are from \llava model. Answer choices in VizWiz examples (shown in gray) are generated negatives from the ground truth, which are not shown to the VLM model when generating the predictions and explanations and are only provided to compute the contrastiveness (shown in gray).}
\label{tab:example-model-explanations}
\end{table*}

\section{VizWiz Dataset Limitations and Negative Discriminability}
\label{sec:apx_vizwiz_negatives}

We observed that for VizWiz, particularly with \gpt, the discriminability scores for our proposed metrics (and baselines) are often low or negative (see \Cref{tab:disc}). A negative discriminability score implies that the model assigns higher quality scores to ``incorrect'' predictions than to ``correct'' ones. 

Upon qualitative inspection, we found this phenomenon is largely driven by the ``Ground Truth Problem'' inherent in the VizWiz dataset. The ground truth answers are often narrow or specific, causing valid model predictions to be labeled as incorrect. \Cref{tab:vizwiz_negatives_examples} shows three representative examples:

\begin{table*}[htbp]
\centering
\resizebox{\linewidth}{!}{
    \begin{tabular}{%
      p{2cm} 
      p{3cm} 
      p{2.5cm} 
      p{1.5cm} 
      p{2cm} 
      p{6cm} 
      p{1cm} 
      p{1cm} 
    }
    \toprule
    \textbf{Question} &
    \textbf{Image} &
    \textbf{Ground Truth List} &
    \textbf{Model} &
    \textbf{Prediction} &
    \textbf{Explanation} &
    \textbf{VF} &
    \textbf{Contr.} \\
    \midrule
    What does the screen say? &
    \includegraphics[width=\linewidth, valign=t]{figs/VQA_examples/VizWiz-127.jpg} &
    windows 7 home premium, windows, windows home premium &
    \gpt &
    set up windows &
    The screen displays the setup process for Windows 7 Home Premium. It prompts the user to choose a user name for their account and name their computer to distinguish it on the network. The text on the screen includes:
- "Set Up Windows"
- "Windows 7 Home Premium"
- "Choose a user name for your account and name your computer to distinguish it on the network."
- "Type a user name (for example, John):"
- "Type a computer name:"
- "PC"
- "Copyright © 2009 Microsoft Corporation. All rights reserved." &
    1.000 &
    0.585 \\
    
    What is this? &
    \includegraphics[width=\linewidth, valign=t]{figs/VQA_examples/VizWiz-750.jpg} &
    tall buildings, buildings, city landscape, downtown city, city buildings, skyscrapers, city&
    \gpt &
    downtown Toronto &
    The image shows a cityscape with several tall buildings, one of which has the TD (Toronto-Dominion) logo prominently displayed. This suggests that the location is likely in downtown Toronto, where TD Bank has a significant presence. The mix of modern and older architecture is also characteristic of downtown Toronto. Therefore, the answer is indeed downtown Toronto. &
    1.000 &
    0.659 \\

    What's this? &
    \includegraphics[width=\linewidth, valign=t]{figs/VQA_examples/VizWiz-687.jpg} &
    room, half computer screen, monitor, laptop, computer screen, computer monitor, tv &
    \qwen &
    keyboard &
    The image shows a piece of electronic equipment that appears to be an electronic keyboard or synthesizer. It has keys similar to a piano and is positioned next to what looks like a computer monitor, suggesting it might be used for music production or performance. The lighting is dim, and the image quality is low, making it difficult to discern finer details, but the shape and layout of the keys are consistent with a keyboard. &
    1.000 &
    0.999 \\
    \bottomrule
    \end{tabular}
}
\caption{Examples of ``Incorrect'' VizWiz predictions that received high Explanation Quality scores. In these cases, the model provides a semantically valid or more specific answer than the ground truth, leading to a ``False Negative'' error judgment despite high Visual Fidelity and Contrastiveness. }
\label{tab:vizwiz_negatives_examples}
\end{table*}

\begin{itemize}
    \item \textbf{Setup Windows Example}: The ground truth includes ``windows 7 home premium'', but the \gpt model predicts ``set up windows'', which accurately describes the visible title. This is marked incorrect, yet receives high quality scores.
    \item \textbf{Toronto Example}: The ground truth is generic (``tall buildings''), but the \gpt model correctly identifies ``downtown Toronto''. This specific identification is penalized, again resulting in high quality scores for a ``wrong'' answer.
    \item \textbf{Keyboard Example}: The ground truth lists various terms for a screen (e.g., ``monitor'', ``laptop''), but \qwen predicts ``keyboard'', identifying the other prominent object in the blurry image. The explanation correctly describes the keyboard's features, leading to high fidelity despite the answer mismatch.
\end{itemize}

To quantify this, an expert annotator manually inspected 20 randomly sampled instances where the model prediction was marked ``incorrect'' but achieved a Visual Fidelity score $>0.9$. We found that for \gpt, 60\% (12/20) of these cases were actually valid answers rejected by the ground truth (e.g., specific vs generic). This rate was lower for \qwen (25\%, 5/20) and \llava (15\%, 3/20), confirming that \gpt's ``hallucinations'' on VizWiz are frequently artifacts of valid but mismatched reasoning.

Because \gpt often provides these ``over-answering'' or semantically valid but lexically distinct responses, it accumulates a large number of high-quality ``incorrect'' samples, which inverts the discriminability metric. This effect is less pronounced in \qwen and least in \llava, which tend to generate shorter, simpler answers closer to the crowd-sourced ground truths. Thus, the negative discriminability reflects a limitation of the evaluation ground truth rather than a failure of the explanation quality metrics.

\section{Reliability and Validity of the Trust Cues}
\label{sec:verifier_reliability}
\subsection{Validation on Agreement with Human Judgements}
\label{subsec:internal_sanity_check}

To validate the reliability of our automated Visual Fidelity and Contrastiveness qualities, we conducted an internal sanity check on a held‐out sample. We randomly selected 100 examples from our evaluation sets and had two expert annotators independently assess each instance. The first annotator is one of the authors of this paper, who has direct experience in VLM evaluation; the second is an independent expert who also has experience with VLM evaluation but is not familiar with this paper or the dataset. The instruction provided to the expert annotators is shown at \Cref{tab:expert_anno_instructions}.

We binarized our scores at a threshold of $0.5$ (scores $\ge 0.5$ map to $1$, else $0$) and computed Cohen’s $\kappa$ both between the two annotators and between each annotator and the automated scores. Inter-annotator agreement between the two experts is
\[
\kappa_{\mathrm{VF}} = 0.68 \quad\text{and}\quad
\kappa_{\mathrm{Contr.}} = 0.46,
\]
confirming that the task is well-defined. Agreement with the automated scores, reported as (Annotator 1 / Annotator 2), is
\[
\kappa_{\mathrm{VF}} = 0.51 \,/\, 0.49 \quad\text{and}\quad
\kappa_{\mathrm{Contr.}} = 0.44 \,/\, 0.39,
\]
indicating moderate and consistent agreement across annotators and between annotators and automated qualities.

\subsection{Accuracy of LLM-generated verification questions and answers}
\label{sec:llm_qna_audit}
To quantify hallucination risk in the VF pipeline, we sampled 50 explanations each from A-OKVQA and VizWiz (total 210 verification questions produced by $m_{\mathrm{QGen}}$ and corresponding answers from $m_{\mathrm{Verif}}$). Three expert annotators judged (i) whether each generated verification question was correctly grounded in the rationale and image, and (ii) whether the verifier’s answer was correct for questions deemed evaluable from the image. We observed that $95.2\%$ of generated questions were correctly grounded and $95.0\%$ of verifier answers were correct. These results support using LLM-as-judge as a cost-efficient approximation to the ideal ground-truth VF scorer.

\subsection{Verifier robustness for Visual Fidelity}
\label{sec:verifier_robustness}
We use \gpt by default for both question generation ($m_{\mathrm{QGen}}$) and verification ($m_{\mathrm{Verif}}$) to isolate pipeline effects. To assess robustness to verifier choice, we swap in smaller open models (\gemma, \qwen) and report changes in Discriminability (Disc; higher is better) and Expected Calibration Error (ECE; lower is better) on held-out slices.

\begin{table*}[!htbp]
\centering
\resizebox{\linewidth}{!}{
\begin{tabular}{l l c c c c}
\toprule
Dataset & Model & Accuracy & Disc VF(GPT) & Disc VF(Gemma) & Disc VF(Qwen) \\
\midrule
A-OKVQA & \llava & $ 0.696 $ & $ 0.181^{***}$ & $ 0.151^{***}$ & $ 0.247^{***}$ \\
A-OKVQA & \qwen & $ 0.848 $ & $ 0.080^{***}$ & $ 0.096^{**\phantom{*}}$     & $ 0.131^{***}$ \\
A-OKVQA & \gpt & $ 0.910 $ & $ 0.085^{***}$ & $ 0.044^{\phantom{***}} $                    & $ 0.156^{***}$ \\
VizWiz   & \llava & $ 0.557 $ & $ 0.240^{***}$ & $ 0.098^{***}$ & $ 0.193^{***}$ \\
VizWiz   & \qwen & $ 0.838 $ & $ 0.042^{*\phantom{**}}$         & $ 0.024^{\phantom{***}} $                    & $ 0.092^{***}$ \\
VizWiz   & \gpt & $ 0.840 $ & $ 0.024^{\phantom{***}} $                    & $-0.048^{\phantom{***}}\phantom{-}$                 & $ 0.104^{**\phantom{*}}$     \\
\midrule
Average  &                         & $ 0.781 $ & $ 0.109^{\phantom{***}} $ & $ 0.061^{\phantom{***}} $ & $ 0.154^{\phantom{***}} $ \\
\bottomrule
\end{tabular}
}
\caption{Discriminability of VF scores across verifier models (higher is better). Significance: ${}^*$ $p{<}.05$, ${}^{**}$ $p{<}.01$, ${}^{***}$ $p{<}.001$.}
\label{tab:verifier_disc}
\end{table*}
\begin{table*}[!htbp]
\centering
\resizebox{\linewidth}{!}{
\begin{tabular}{l l c c c c}
\toprule
Dataset & Model & Accuracy & ECE VF(GPT) & ECE VF(Gemma) & ECE VF(Qwen) \\
\midrule
A-OKVQA & \llava          & $ 0.696 $ & $ 0.207 $ & $ 0.156 $ & $ 0.181 $ \\
A-OKVQA & \qwen          & $ 0.848 $ & $ 0.137 $ & $ 0.174 $ & $ 0.177 $ \\
A-OKVQA & \gpt & $ 0.910 $ & $ 0.099 $ & $ 0.209 $ & $ 0.173 $ \\
VizWiz   & \llava          & $ 0.557 $ & $ 0.271 $ & $ 0.303 $ & $ 0.310 $ \\
VizWiz   & \qwen          & $ 0.838 $ & $ 0.136 $ & $ 0.134 $ & $ 0.161 $ \\
VizWiz   & \gpt & $ 0.840 $ & $ 0.160 $ & $ 0.205 $ & $ 0.174 $ \\
\midrule
Average  &                         & $ 0.781 $ & $ 0.168 $ & $ 0.197 $ & $ 0.196 $ \\
\bottomrule
\end{tabular}
}
\caption{Expected Calibration Error (ECE) of VF scores across verifier models (lower is better).}
\label{tab:verifier_ece}
\end{table*}

As shown in \Cref{tab:verifier_disc} and \Cref{tab:verifier_ece}, across A-OKVQA and VizWiz, all three verifiers yield positive discriminability on average (\gpt: 0.109; \gemma: 0.061; \qwen: 0.154), and ECE remains comparably low (0.168/0.197/0.196 respectively). Thus, VF continues to separate correct from incorrect predictions and stays well-calibrated even with using smaller open models as verifiers.

\begin{table*}[t]
  \centering
  \begin{tabular}{p{0.16\linewidth} p{0.8\linewidth}}
    \toprule
    \textbf{Criterion} & \textbf{Instruction} \\
    \midrule
    Visual Fidelity & 
        In this task, you’ll view an image, and a question about the image. You will then see an answer to the question given by an AI model, along with an explanation.\newline
        Your job is to evaluate whether the details in the explanation are consistent with the image that is shown.\newline
        You'll select one of these choices:\newline
        0: The explanation does not mention any details / elements that are directly visible in the image (apart from the prediction itself). Or the explanation mentions details about the image, but one or more of those details are incorrect (they contradict what is visible in the image).\newline
        1: The explanation mentions details about the image, and all the details are consistent with the image.\newline
        IMPORTANT NOTE: Your job is NOT to check the correctness of the AI model's answer. It could be that the answer or the logic in the response is incorrect, but the explanation talks about something that is directly in the image. It can also be that the answer is correct, but the explanation does not refer to the image or presents inconsistent details!\newline
        IMPORTANT: See examples from this \href{https://forms.gle/nvpcQ7fpLoFqsLiJ9}{form (https://forms.gle/nvpcQ7fpLoFqsLiJ9)} before proceeding!\\
    \midrule
    Contrastiveness   & In this task, you'll view a question about an image (without seeing the image). You will also see an answer from an AI model, along with an explanation (the model has access to the image). Your job is to evaluate whether the explanation meets these two key qualities:
    \begin{itemize}[nosep, nolistsep]
        \item You check if the explanation is consistent with the predicted answer.
        \item You can ask yourself this question: ``Does the explanation provide evidence that matches with the answer it gives?''
    \end{itemize}
    You check if the explanation covers enough details to reject all other possible answers.
    You can ask yourself this question: ``Does the explanation eliminate all other answers with proper justifications?''
    By eliminate, we mean that the explanation should provide a strong argument for the selection answer, or strong counter arguments for the other options.
    You'll decide whether the explanation has these qualities. \newline
    IMPORTANT: See examples from this \href{https://forms.gle/HYp6N44TpsjAoxHa6}{form (https://forms.gle/HYp6N44TpsjAoxHa6)} before proceeding!\\
    \bottomrule
  \end{tabular}
  \caption{Instructions Provided to the Expert Annotator}
  \label{tab:expert_anno_instructions}
\end{table*}

\section{User Studies}
\label{sec:apx_user_studies}

\subsection{Annotator filtering}
\label{subsec:annotator_filering}
\begin{table*}[t]
    \centering
    
    \begin{tabularx}{\textwidth}{%
        >{\raggedright\arraybackslash}p{0.4\textwidth} | 
        >{\centering\arraybackslash}p{0.56\textwidth} 
        }
        \toprule
        Config & Criteria\\
        \midrule
        Location & United States \\
        Current Country of Residence & 
        United States
        \\
        Primary Language & English\\
        Approval Rate & 98-100 \\
        Number of previous submissions & 1000-10000\\
        Highest education level completed & Undergraduate degree (BA/BSc/other), Graduate degree (MA/MSc/MPhil/other), Doctorate degree (PhD/other)\\
        Exclude participants from other studies & Prohibit any user who took part in a different setting\\
        \bottomrule
    \end{tabularx}
    \caption{Prolific annotator filtering}
    \label{tab:annotator_filtering}
\end{table*}

To ensure high-quality annotations, we recruited annotators on Prolific using the criteria listed in \Cref{tab:annotator_filtering}. All participants identities were fully anonymized. All annotators provided informed consent through Prolific’s platform, which clearly described the task purpose, data use, and compensation prior to participation. 

This study protocol was reviewed and approved by the University of Southern California's Institutional Review Board.

All participants were required to be native English speakers residing in the United States, with approval rates between 98–100\% and at least 1000 prior submissions. We further restricted enrollment to individuals holding at least an undergraduate degree. To prevent contamination across experimental conditions, each participant was confined to a single “setting” (i.e., one quality-type configuration).

\subsection{Dataset Splits and User Study Design}
\label{subsec:create_subset_split}

\paragraph{Metric Evaluation Sets (500 instances)}
For the quality evaluations detailed in \Cref{sec:metric_experiments}, we selected three 500-instance sets. From A-OKVQA, we used the first 500 examples from the official validation split. From VizWiz, we selected the first 500 validation examples where the majority human annotation was not ``unanswerable'' and which contained no NSFW content, as filtered by \gpt. From MMMU-Pro, we used the first 500 examples with single-image contexts from the \textit{standard (10 options)} set.

\paragraph{Human Study Subsets (100 instances)}
We conducted a between-subjects user study where participants predicted VLM correctness based on varying levels of explanation quality scores. To facilitate this, we derived a subset of 100 questions each from the A-OKVQA and VizWiz evaluation sets described above.

These subsets were balanced to contain 50 correct and 50 incorrect predictions by \llava. To ensure the subsets remained representative of the larger distribution, we generated 50 random candidate subsets and selected the one with the lowest Expected Calibration Error (ECE) relative to the full 500-instance set. Specifically, for A-OKVQA, we minimized the average ECE across Visual Fidelity and Contrastiveness; for VizWiz, we minimized the ECE for Visual Fidelity (as Contrastiveness scores were unavailable during the subset creation phase). As shown in \Cref{tab:ece_metrics_aokvqa} and \Cref{tab:ece_metrics_vizwiz}, this selection maintains a distribution of quality scores consistent with the full set from \Cref{subsec:vl-reasoning-task}.

\paragraph{Study Protocol}
In the study, each question was annotated by three unique participants, yielding 30 participants and 300 annotations per dataset per condition. Each participant completed a batch of 10 questions, balanced to include five instances where the model was correct and five where it was incorrect; participants were not informed of this distribution. Users were randomly assigned to a single study condition (control or a treatment variant) and did not participate in multiple conditions. In total, we collect 300 human annotations per dataset and condition (3 users × 100 questions), capturing user reliance behavior under varying explanation and quality signal conditions.

\begin{table*}[ht]
\centering
\resizebox{\linewidth}{!}{%
\begin{tabular}{lc|ccccc|ccc}
\toprule
Set & Accuracy &  VF &  Contr. &  Prod(VF, Contr.) &  Min(VF, Contr.) &  Avg(VF, Contr.) &  Support &  Informative &  Plausibility \\
\midrule
Full (500)  & 0.696 & 0.207 & 0.176 & 0.164 & 0.147 & 0.109 & 0.288 & 0.332 & 0.162 \\
Subset (100)  & 0.500 & 0.270 & 0.237 & 0.133 & 0.130 & 0.227 & 0.372 & 0.490 & 0.075 \\
\bottomrule
\end{tabular}%
}
\caption{Accuracy and Expected Calibration Error (ECE) for different qualities on the \textbf{A-OKVQA} dataset. ``Subset'' rows correspond to the 100-question user study subsets (50 correct / 50 incorrect), while ``Full'' rows represent the 500-instance evaluation set.}
\label{tab:ece_metrics_aokvqa}

\bigskip

\resizebox{0.55\linewidth}{!}{%
\begin{tabular}{lc|c|ccc}
\toprule
Set & Accuracy &  VF &  Support &  Informative &  Plausibility \\
\midrule
Full (500)  & 0.557 & 0.271 & 0.305 & 0.429 & 0.110 \\
Subset (100)  & 0.500 & 0.236 & 0.379 & 0.390 & 0.071 \\
\bottomrule
\end{tabular}%
}
\caption{Accuracy and Expected Calibration Error (ECE) for different qualities on the \textbf{VizWiz} dataset.}
\label{tab:ece_metrics_vizwiz}
\end{table*}

\subsection{Attention Incentives}
\label{subsec:apx_attention_incentives}

To ensure that participants engaged carefully with each instance's annotation, we combined a per-item timer with small monetary bonuses and penalties.

\subsubsection{Timer Implementation}
\label{subsubsec:timer_implementation}

To ensure that participants spent sufficient time on each stage (and did not simply skim and click through), we imposed a per-item timer in all of our human studies. The user may only start their selection after the timer ends.
For each question, we computed the explanation's ``reading time'' as
\[
\mathrm{reading\_time} = \frac{\#\text{words in explanation}}{238\ \text{words/minute}}
\]
where 238 wpm is the average adult reading speed~\cite{brysbaert2019many}.

We then capped the total display time at
\begin{align*}
\mathrm{reading\_time} + 10\ \text{seconds}
\end{align*}
to cover the question, answers, and any qualites shown.

\subsubsection{Bonus Payments}
\label{subsubsec:apx_bonus_payments}
To further motivate careful reading and discourage guessing, we tied each response to a small bonus bank: correct answers earned $+\$0.10$; incorrect answers incurred $-\$0.10$ (with the bonus bank floored at $\$0$ so it would not harm their base payments); selecting ``I'm unsure based on the information provided'' resulted in no change. Participants were paid their accumulated bonus in addition to the base participation fee of $\$2$ for annotating 10 instances.

\subsection{User Studies Settings \& Examples}
\label{subsec:one_stage_user_studies}
\begin{table*}[ht]
\centering
\small
\begin{tabular}{lp{12cm}}
\toprule
\textbf{Setting}               & \textbf{Description}                             \\
\midrule
Control                     & Show explanation only; no quality scores are shown.       \\
Random Score                  & Random scoring baseline (uniform random distribution from $[0,1]$), shown as a simple confidence score ``AI Confidence that the explanation is correct''.                          \\
Prod(VF, Contr)                & Product of VF and Contrast scores, shown as a simple confidence score ``AI Confidence that the explanation is correct''.               \\
Avg(VF, Contr)                 & Average of VF and Contrast scores, shown as a simple confidence score ``AI Confidence that the explanation is correct''.               \\
VF num.                         & Show Visual Fidelity numeric score  ``AI Confidence that the explanation accurately describes the image details''                      \\
VF desc.                        & Show at most two descriptive sentences converted from visual questions which are verified by verifier VLM $\model{\verif}$, and at most two from questions which are not verified by $\model{\verif}$           \\
Contr. num.                      & Show Contrastiveness score  ``AI Confidence that the explanation rules out the other choices''                  \\
Contr. desc.                     & Show the other answer options $a_j\neq a_0$ s.t. $\textsc{Pr}_{\nli}(P \text{ entails } h_j) \geq 0.5$                 \\
Both num.                   & Display both VF num. and Contr. num. messages                   \\
Both desc.               & Display both VF desc. and Contr. desc. messages                 \\
VF shown as Conf.               & VF score displayed as a simple confidence score ``AI Confidence that the explanation is correct''       \\
Contr shown as Conf.            & Contr. score displayed as a simple confidence score ``AI Confidence that the explanation is correct''\\
Prod shown as VF               & VF$\times$Contr. score presented as a VF score ``AI Confidence that the explanation accurately describes the image details''   \\
Prod shown as Contr.            & Combined VF$\times$Contr. score presented as a Contr. score ``AI Confidence that the explanation rules out the other choices''  \\
\bottomrule
\end{tabular}
\caption{Summary of the 14 one‐step human‐study settings.}
\label{tab:user_study_settings}
\end{table*}

\begin{figure*}[!htbp]
  \centering
  \begin{subfigure}[b]{0.48\linewidth}
    \includegraphics[width=\linewidth, clip]{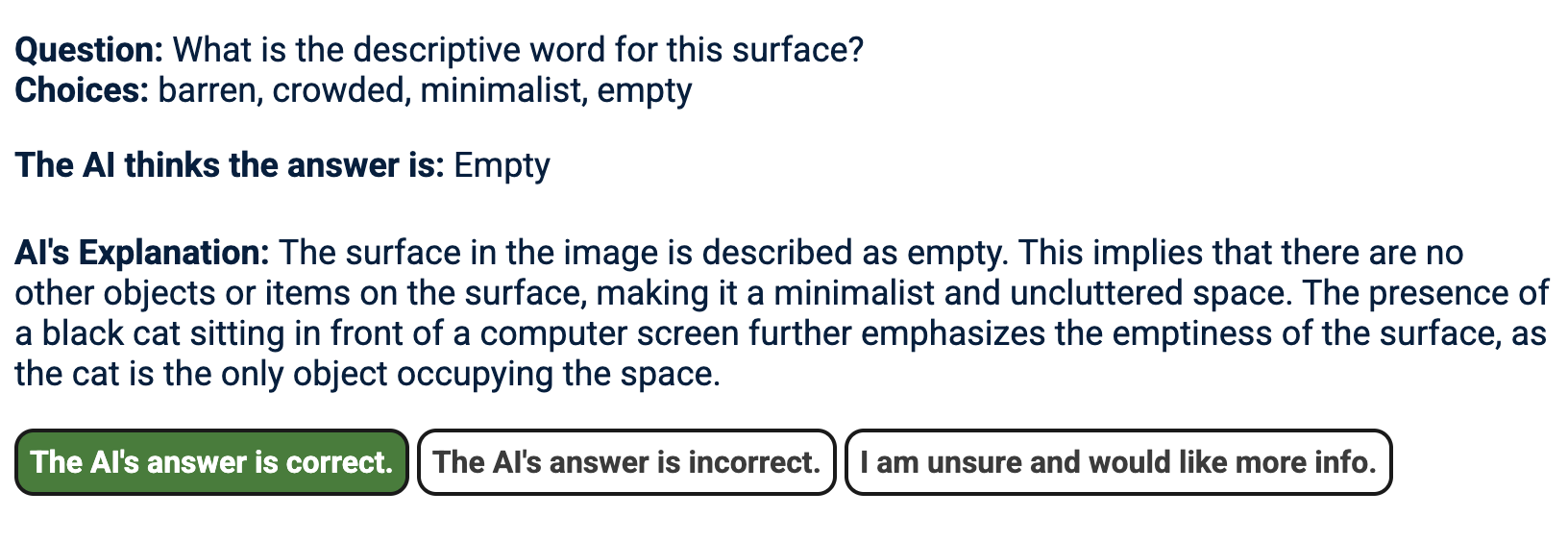}
    \caption{Control (no quality)}\label{sf:combined12}
  \end{subfigure}\hfill
  \begin{subfigure}[b]{0.48\linewidth}
    \includegraphics[width=\linewidth, trim={0 0 0 200}, clip]{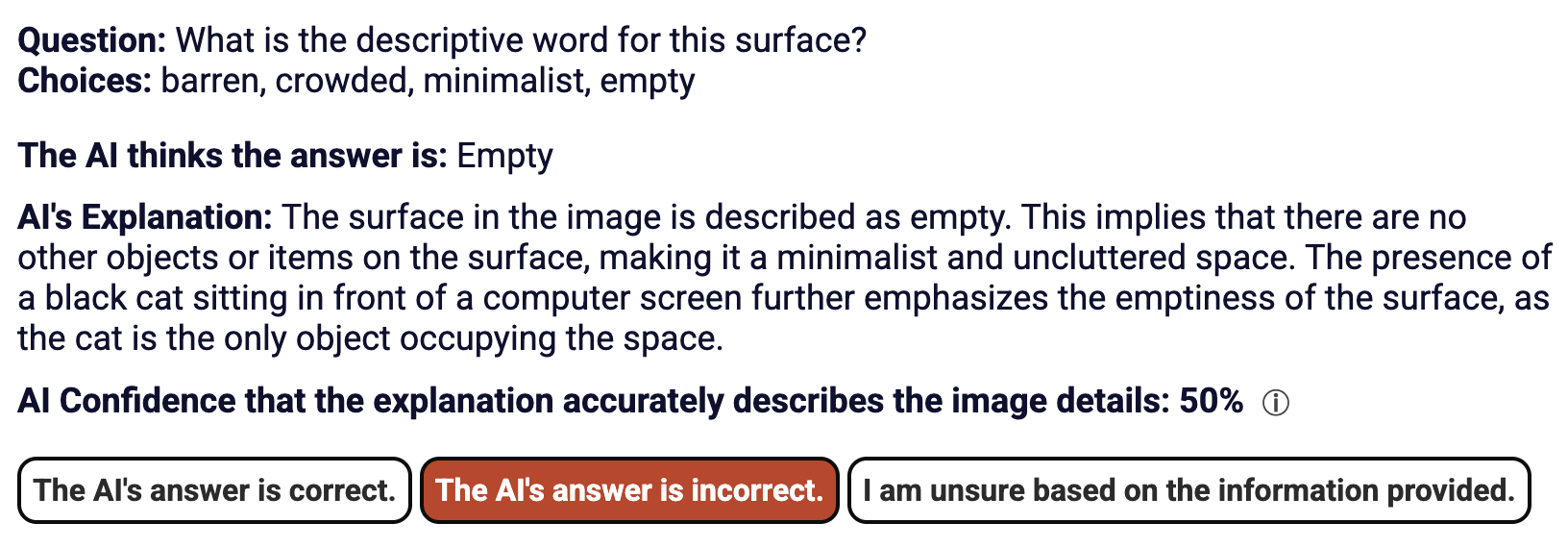}
    \caption{VF – Numeric quality}\label{sf:vf_num}
  \end{subfigure}

  \vspace{1ex}

  \begin{subfigure}[b]{0.48\linewidth}
    \includegraphics[width=\linewidth, trim={0 0 0 200}, clip]{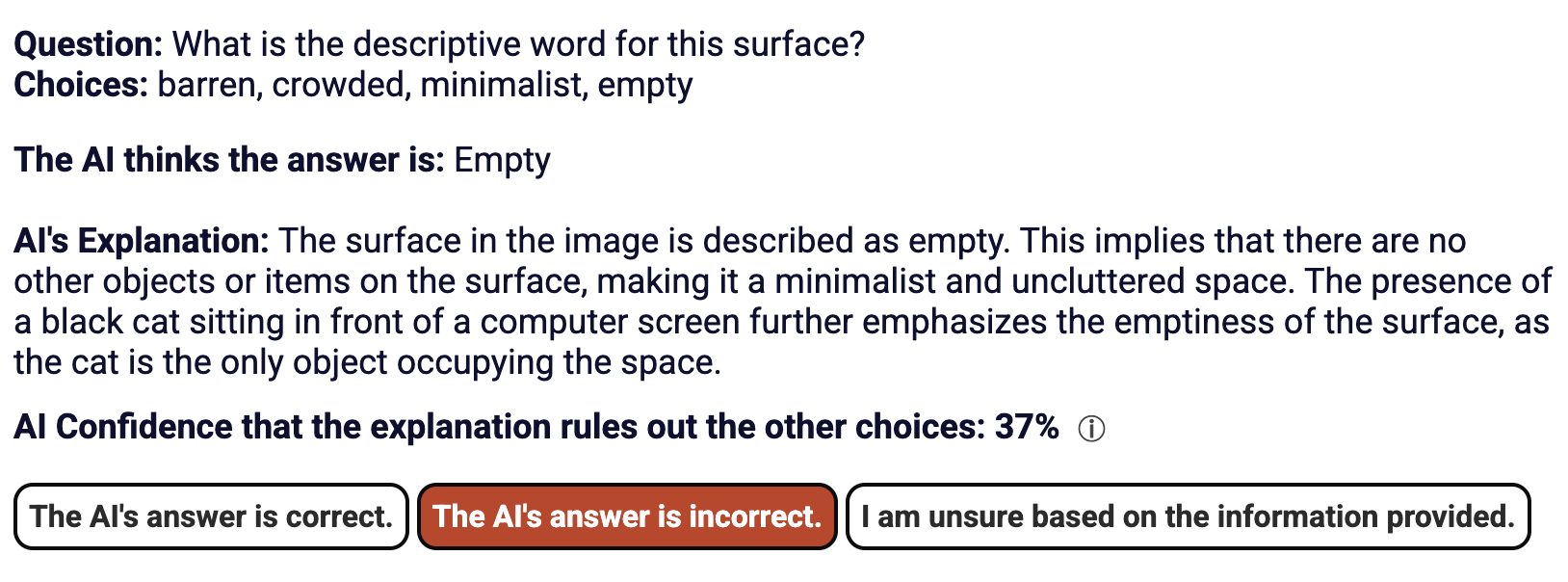}
    \caption{Contr. – Numeric quality}\label{sf:ctr_num}
  \end{subfigure}\hfill
  \begin{subfigure}[b]{0.48\linewidth}
    \includegraphics[width=\linewidth, trim={0 0 0 200}, clip]{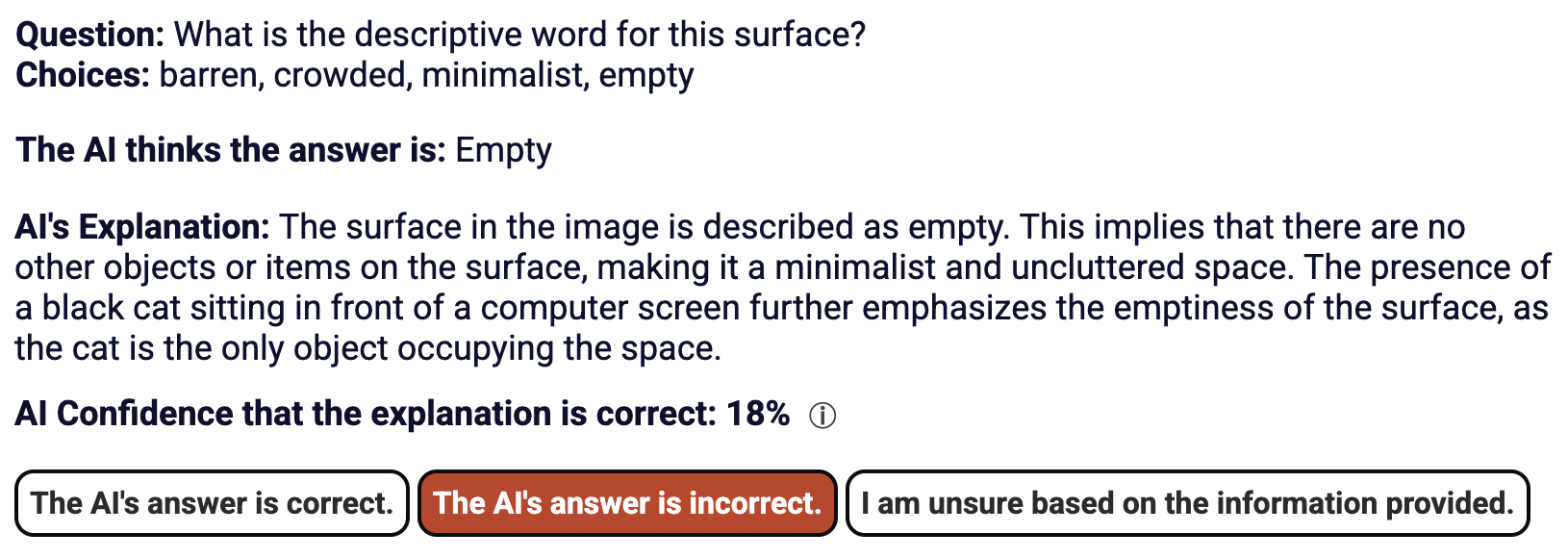}
    \caption{Simple confidence score – Numeric quality (used for Prod(VF, Contr.) and Random Score)}\label{sf:single_num}
  \end{subfigure}

  \vspace{1ex}

  \begin{subfigure}[b]{0.48\linewidth}
    \includegraphics[width=\linewidth, trim={0 0 0 200}, clip]{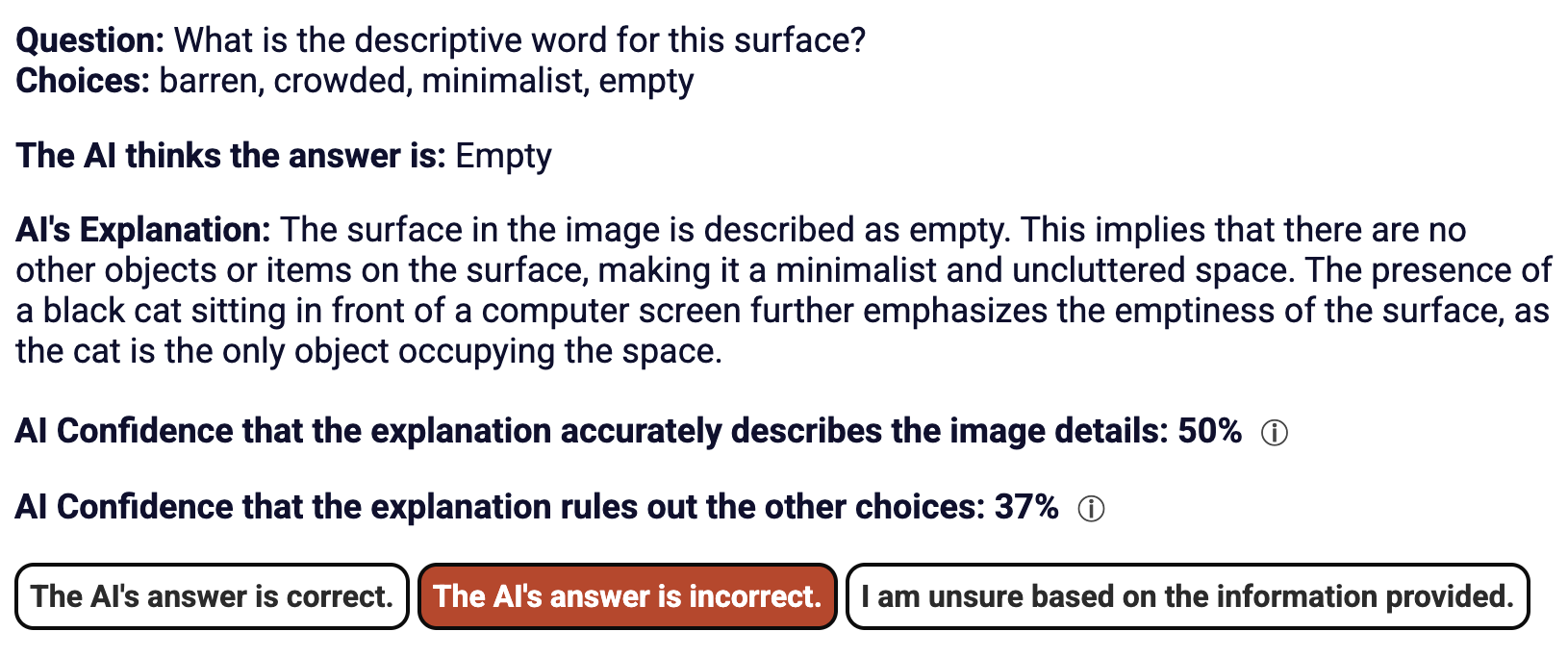}
    \caption{Both VF and Contr. – Numeric quality}\label{sf:vf_contr_num}
  \end{subfigure}\hfill
  \begin{subfigure}[b]{0.48\linewidth}
    \includegraphics[width=\linewidth, trim={0 0 0 200}, clip]{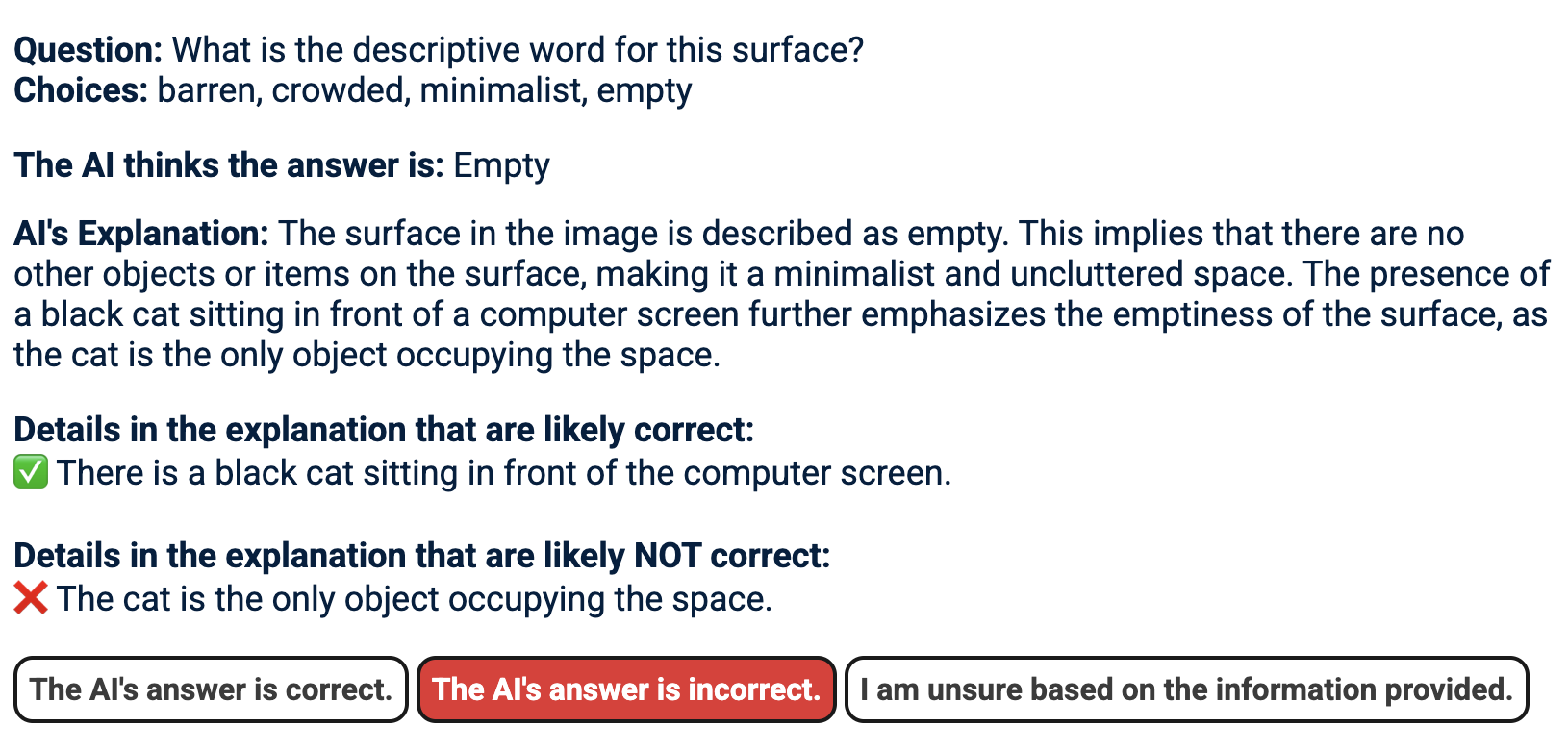}
    \caption{VF – Descriptive quality}\label{sf:vf_desc}
  \end{subfigure}

  \vspace{1ex}

  \begin{subfigure}[b]{0.48\linewidth}
    \includegraphics[width=\linewidth, trim={0 0 0 200}, clip]{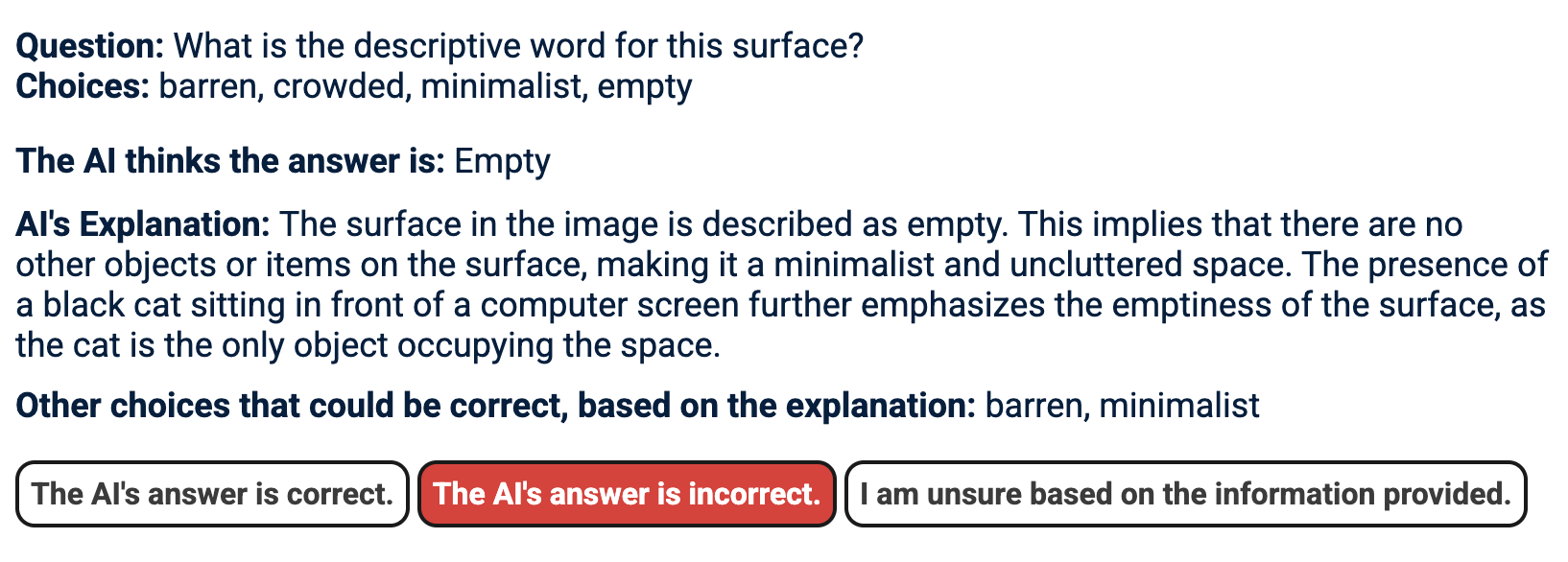}
    \caption{Contr. – Descriptive quality}\label{sf:ctr_desc}
  \end{subfigure}\hfill
  \begin{subfigure}[b]{0.48\linewidth}
    \includegraphics[width=\linewidth, trim={0 0 0 200}, clip]{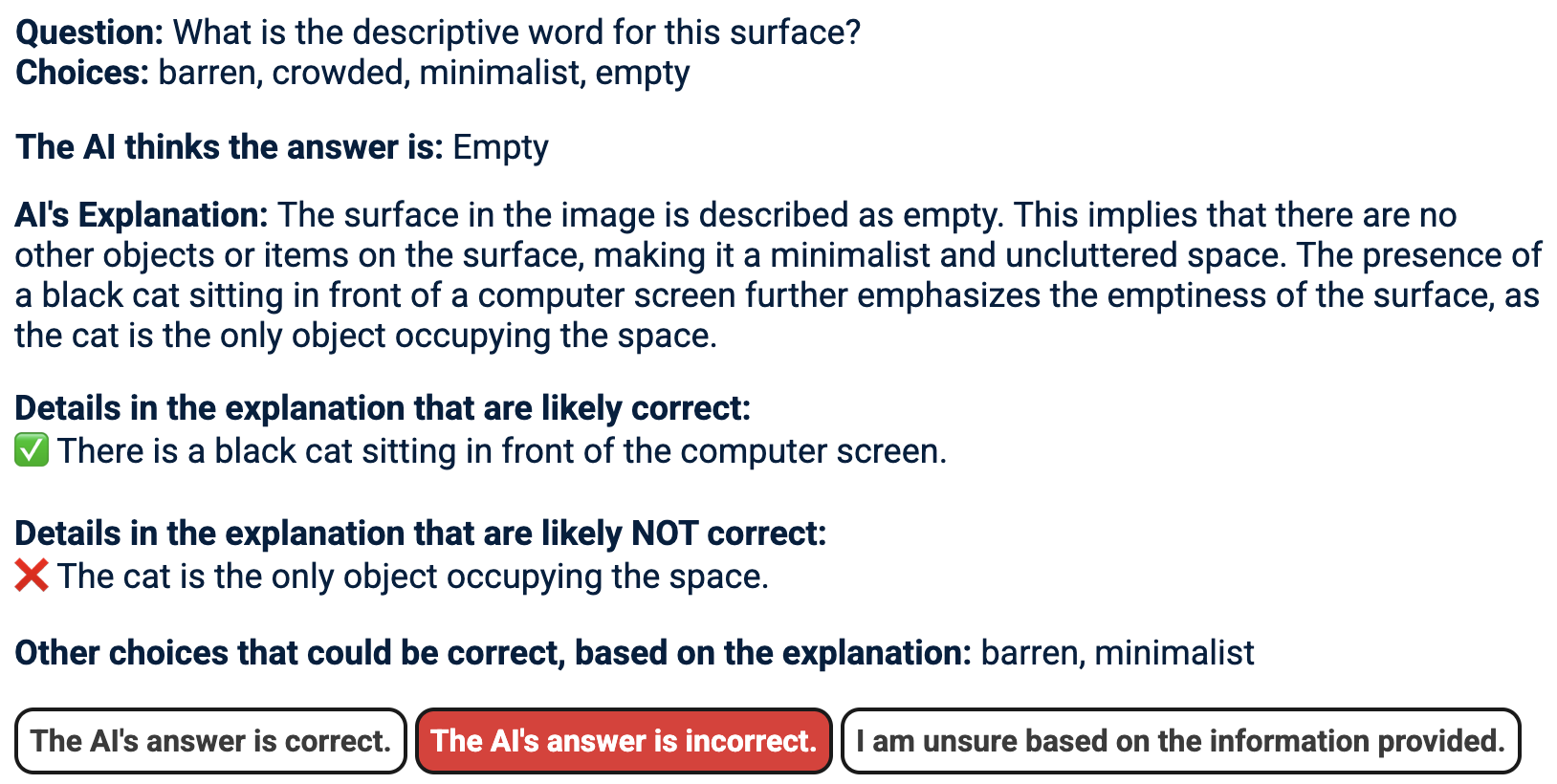}
    \caption{Both VF and Contr. – Descriptive quality}\label{sf:vf_ctr_desc}
  \end{subfigure}

  \caption{Explanation quality messages for each instruction condition. Subfigure \subref{sf:combined12} is the baseline with no quality displayed; the next three (\subref{sf:vf_num}–\subref{sf:vf_contr_num}) show numeric qualities; the remaining four (\subref{sf:vf_desc}–\subref{sf:vf_ctr_desc}) show descriptive ones.}
  \label{fig:all_setting_examples}
\end{figure*}

\begin{table*}[!htbp]
\centering

\resizebox{\textwidth}{!}{%
\begin{tabular}{llp{2.2cm}p{2.2cm}p{2.2cm}p{2.2cm}p{2.2cm}}
\toprule
Dataset& Setting  & Unsure Rate& Not Unsure Accept Rate & Not Unsure Accuracy & False Accept Rate & False Reject Rate \\
\midrule
\multirow{14}{*}{A-OKVQA}
& Control &  08.7\% ± 1.6\% & 76.3\% ± 2.6\%& 59.1\% ± 3.0\%& 60.7\% ± 2.7\%  & 14.0\% ± 1.5\%\\
& Random Score& 14.3\% ± 2.0\% & 70.0\% ± 2.9\%& 59.9\% ± 3.1\%& 50.7\% ± 2.5\%  & 18.0\% ± 1.7\%\\
& Prod(VF, Contr.) &  09.3\% ± 1.7\% & 69.9\% ± 2.8\%& 70.2\% ± 2.8\%& 45.3\% ± 2.4\%  &  08.7\% ± 1.2\%\\
& Avg(VF, Contr.)  &  07.3\% ± 1.5\% & 74.8\% ± 2.6\%& 66.2\% ± 2.8\%& 52.7\% ± 2.5\%  & 10.0\% ± 1.3\%\\
& VF num. &  07.3\% ± 1.5\% & 76.3\% ± 2.6\%& 62.9\% ± 2.9\%& 58.7\% ± 2.6\%  & 10.0\% ± 1.3\%\\
& VF desc. &  09.7\% ± 1.7\% & 69.4\% ± 2.8\%& 68.6\% ± 2.8\%& 45.3\% ± 2.4\%  & 11.3\% ± 1.3\%\\
& Contr. num. &  09.0\% ± 1.7\% & 77.7\% ± 2.5\%& 64.1\% ± 2.9\%& 56.7\% ± 2.6\%  &  08.7\% ± 1.2\%\\
& Contr. desc. &  06.0\% ± 1.4\% & 71.6\% ± 2.7\%& 61.3\% ± 2.9\%& 56.7\% ± 2.6\%  & 16.0\% ± 1.6\%\\
& Both num. & 10.7\% ± 1.8\% & 72.8\% ± 2.7\%& 66.4\% ± 2.9\%& 48.7\% ± 2.5\%  & 11.3\% ± 1.3\%\\
& Both desc.&  08.7\% ± 1.6\% & 68.2\% ± 2.8\%& 67.2\% ± 2.8\%& 46.0\% ± 2.4\%  & 14.0\% ± 1.5\%\\
& VF shown as Conf.&  07.0\% ± 1.5\% & 77.8\% ± 2.5\%& 61.3\% ± 2.9\%& 60.0\% ± 2.7\%  & 12.0\% ± 1.4\%\\
& Contr. shown as Conf.&  08.3\% ± 1.6\% & 74.5\% ± 2.6\%& 69.1\% ± 2.8\%& 50.0\% ± 2.5\%  &  06.7\% ± 1.0\%\\
& Prod shown as VF&  07.7\% ± 1.5\% & 65.7\% ± 2.9\%& 67.5\% ± 2.8\%& 44.4\% ± 2.4\%  & 16.0\% ± 1.6\%\\
& Prod shown as Contr.&  08.0\% ± 1.6\% & 70.3\% ± 2.8\%& 68.5\% ± 2.8\%& 44.7\% ± 2.4\%  & 13.3\% ± 1.4\%\\
\midrule
\multirow{3}{*}{VizWiz}  & Control  &  07.3\% ± 1.5\% & 70.5\% ± 2.7\%& 59.4\% ± 3.0\%& 55.3\% ± 2.6\%  & 20.0\% ± 1.7\%\\
& VF num. &  09.3\% ± 1.7\% & 65.1\% ± 2.9\%& 64.7\% ± 2.9\%& 44.7\% ± 2.4\%  & 19.3\% ± 1.7\%\\
& VF desc. & 10.3\% ± 1.8\% & 67.3\% ± 2.9\%& 69.1\% ± 2.8\%& 42.0\% ± 2.4\%  & 13.3\% ± 1.4\%\\
\bottomrule
\end{tabular}
}
\caption{User-study results on A-OKVQA and VizWiz in the one-stage setting. Each row received 300 annotations (100 questions $\times$ 3 annotations per question).}
\label{tab:one_stage_results}
\end{table*}

In the main user study, we present 14 settings in total. \Cref{tab:user_study_settings} summarizes all 14 different settings and \Cref{fig:all_setting_examples} shows examples of all 8 different categories of messages presented to users.
\Cref{tab:one_stage_results} contains the main user-study results of user behavior patterns.

\subsection{Supplementary Human Studies}
\label{subsec:three_stages_user_studies}
In addition to the study described in \Cref{subsec:userstudy_experiments} and \Cref{subsec:one_stage_user_studies}, we ran a supplementary user study to evaluate how explanations and qualities affect users' trust in model predictions. Each of the 10 questions was presented in three successive stages; after each stage, participants indicated whether they believed the model's answer was correct or were unsure:

\begin{enumerate}
\item Answer Only: Participants viewed only the question, answer choices (if available), and model prediction.

\item With Explanation: Participants were provided with AI-generated rationales alongside predictions.

\item With Explanation + Quality: Qualities (varied from our experiment settings, e.g. Visual Fidelity and Contrastiveness) were displayed alongside explanations.
\end{enumerate}

This three-stage design of the user study enables us to track how users’ confidence in the model’s correctness evolves as they receive additional information. 

\paragraph{Timed Stages in Supplementary Human Studies}
To standardize attention across conditions, we also enforced a stage-specific timer, where the users can only make their selections at a stage after the timer at that stage ends:
\begin{enumerate}
    \item Answer Only: fixed 5 seconds
    \item With Explanation: explanation reading time (words / 238 wpm) (roughly 10--40 seconds)
    \item Explanation + Quality: fixed 5 seconds
\end{enumerate}

\paragraph{Bonus Payments in Supplementary Human Studies}
Participants were paid a \$2 base fee and could earn up to \$1 in performance‐based bonuses, which were awarded only during Stage 3 (Explanation + Quality; see \Cref{subsubsec:apx_bonus_payments}).

\begin{table*}[htbp]
\centering
\scriptsize
\resizebox{\textwidth}{!}{%
\begin{tabular}{@{}llp{16mm}r*{2}{c}*{2}{c}*{2}{c}@{}}
\toprule
\multirow{2}{*}{Dataset} & \multirow{2}{*}{VLM} & \multirow{2}{*}{User study setting} & \multirow{2}{*}{\#Ann.} 
  & \multicolumn{2}{c}{After Stage 1} 
  & \multicolumn{2}{c}{After Stage 2} 
  & \multicolumn{2}{c}{After Stage 3} \\
\cmidrule(lr){5-6} \cmidrule(lr){7-8} \cmidrule(lr){9-10}
 & & & 
  & Unsure Rate & User Accuracy 
  & Unsure Rate & User Accuracy
  & Unsure Rate & User Accuracy \\
\midrule
\multirow{7}{*}{A‑OKVQA} 
  & \multirow{7}{*}{LLaVA} 
    & VF num.                & 300 
      & $65.3\%\pm2.8\%$ & $67.3\%\pm4.6\%$ 
      & $19.0\%\pm2.3\%$ & $63.4\%\pm3.1\%$ 
      & $5.0\%\pm1.3\%$  & $63.5\%\pm2.9\%$ \\

  &   & Contr num.            & 300 
      & $65.7\%\pm2.7\%$ & $58.3\%\pm4.9\%$ 
      & $17.7\%\pm2.2\%$ & $60.7\%\pm3.1\%$ 
      & $6.3\%\pm1.4\%$  & $61.2\%\pm2.9\%$ \\

  &   & Both num.       & 300 
      & $77.3\%\pm2.4\%$ & $58.8\%\pm6.0\%$ 
      & $38.0\%\pm2.8\%$ & $54.8\%\pm3.7\%$ 
      & $9.7\%\pm1.7\%$  & $64.2\%\pm2.9\%$ \\

  &   & Avg(VF, Contr.)                  & 300 
      & $86.0\%\pm2.0\%$ & $54.8\%\pm7.8\%$ 
      & $35.3\%\pm2.8\%$ & $63.9\%\pm3.5\%$ 
      & $6.0\%\pm1.4\%$  & $63.5\%\pm2.9\%$ \\

  &   & VF desc.            & 300 
      & $67.3\%\pm2.7\%$ & $59.2\%\pm5.0\%$ 
      & $19.7\%\pm2.3\%$ & $59.3\%\pm3.2\%$ 
      & $3.7\%\pm1.1\%$  & $62.3\%\pm2.9\%$ \\

  &   & Contr desc.         & 300 
      & $63.0\%\pm2.8\%$ & $64.9\%\pm4.6\%$ 
      & $17.0\%\pm2.2\%$ & $63.1\%\pm3.1\%$ 
      & $8.7\%\pm1.6\%$  & $59.5\%\pm3.0\%$ \\

  &   & Both desc.    & 290 
      & $70.7\%\pm2.7\%$ & $47.1\%\pm5.4\%$ 
      & $23.8\%\pm2.5\%$ & $57.9\%\pm3.3\%$ 
      & $8.3\%\pm1.6\%$  & $64.7\%\pm2.9\%$ \\
\midrule
\multirow{2}{*}{VizWiz} 
  & \multirow{2}{*}{Qwen2.5} 
    & VF num.                & 300 
      & $57.7\%\pm2.9\%$ & $59.8\%\pm4.4\%$ 
      & $05.7\%\pm1.3\%$  & $65.7\%\pm2.8\%$ 
      & $1.7\%\pm0.7\%$  & $65.1\%\pm2.8\%$ \\

  &   & VF desc.            & 300 
      & $58.7\%\pm2.8\%$ & $66.1\%\pm4.3\%$ 
      & $17.7\%\pm2.2\%$ & $65.2\%\pm3.0\%$ 
      & $4.7\%\pm1.2\%$  & $67.5\%\pm2.8\%$ \\
\bottomrule
\end{tabular}}
\caption{User-study results on A‑OKVQA and VizWiz in the three-stages setting.}
\label{tab:three_stage_study}
\end{table*}

\Cref{tab:three_stage_study} shows that as users progress from seeing only the model’s answer to viewing explanations and then explanations with quality scores, their unsure rate steadily decreases while user accuracy correspondingly increases, with the largest gains observed when descriptive qualities are provided alongside explanations.
These findings support our findings that richer, more interpretable quality signals can meaningfully improve users' trust calibration.

\section{Computational Resources Spent \& Total Cost}
\label{sec:apx_computational_cost}
The complete breakdown of all monetary expenditures across the study is given in \Cref{tab:cost-breakdown}. In brief, the computational expenses include six hours of A100 GPU time for running the vision‐language models, plus OpenAI API calls for generating predicted answers, explanations, and computing qualities. 
Human evaluation costs cover two Prolific studies---a one‐stage study as shown in \Cref{subsec:one_stage_user_studies} and a three‐stage follow‐up from \Cref{subsec:three_stages_user_studies} (including up to \$1.00 in per‐participation bonuses; see \Cref{subsubsec:apx_bonus_payments}).

Altogether, these sum to an overall expense of approximately \$2,827.

\begin{table*}[!htbp]
\centering

\resizebox{\textwidth}{!}{%
\begin{tabular}{@{}p{0.6\linewidth}lr@{}}
\toprule
\textbf{Study \& Step}                   & \textbf{Activity}                        & \textbf{Total Cost} \\ 
\midrule
Predicted answer \& Explanation Generation (\llava \& \qwen)    
& GPU inference (A100, 6h)  & --- \\
Predicted answer \& Explanation Generation (\gpt)
& OpenAI API call & $\approx$\$30 \\
Qualities Computation & OpenAI API call & $\approx$\$120\\
Human Study (One-stage)            & Prolific participant pay & \$1,561\\
Supplementary Human Study (Three-stages)          & Prolific participant pay & \$1,116\\

\midrule
 & & \textbf{$\approx$\$2,827}    \\
\bottomrule
\end{tabular}
}
\caption{Monetary Cost Breakdown}
\label{tab:cost-breakdown}
\end{table*}

\end{document}